%% file: main.tex
\documentclass[nohyperref]{article}

\usepackage{microtype}
\usepackage{graphicx}
\usepackage{subfigure}
\usepackage{booktabs} %

\usepackage[pdftitle={Scalable Deep Gaussian Markov Random Fields for General Graphs}]{hyperref}

\usepackage[accepted]{icml2022}

\usepackage{amsmath}
\usepackage{amssymb}
\usepackage{mathtools}
\usepackage{amsthm}

\usepackage[capitalize,noabbrev]{cleveref}

\theoremstyle{plain}

\theoremstyle{definition}

\theoremstyle{remark}

\usepackage[textsize=tiny]{todonotes}

\let\originalleft\left
\let\originalright\right
\renewcommand{\left}{\mathopen{}\mathclose\bgroup\originalleft}
\renewcommand{\right}{\aftergroup\egroup\originalright}

\usepackage[acronyms, shortcuts]{glossaries}
\glsdisablehyper
\input{macros}

\input{paper_macros}
\input{acronyms}

\usepackage{paralist}
\usepackage{multibib}
\newcites{app}{Appendix References}

\usepackage{xspace}

\usepackage{fancyvrb}
\usepackage{upquote}

\icmltitlerunning{Scalable Deep Gaussian Markov Random Fields for General Graphs}

\begin{document}
\input{cover_page}
\newpage

\twocolumn[
\icmltitle{Scalable Deep Gaussian Markov Random Fields for General Graphs}

\icmlsetsymbol{equal}{*}

\begin{icmlauthorlist}
\icmlauthor{Joel Oskarsson}{liu}
\icmlauthor{Per Sid\'en}{liu,arriver}
\icmlauthor{Fredrik Lindsten}{liu}
\end{icmlauthorlist}

\icmlaffiliation{liu}{Division of Statistics and Machine Learning, Department of Computer and Information Science, Link\"{o}ping University, Link\"{o}ping, Sweden}
\icmlaffiliation{arriver}{Arriver Software AB}

\icmlcorrespondingauthor{Joel Oskarsson}{joel.oskarsson@liu.se}

\icmlkeywords{Machine Learning, ICML, DGMRF, Deep Gaussian Markov Random Field, Graph, Scalable, GNN, Graph Neural Network, Variational Inference, Network, Graph, Graphical Model}

\vskip 0.3in
]

\printAffiliationsAndNotice{}  %

\begin{abstract}
\input{abstract}
\end{abstract}

\section{Introduction}
\label{sec:introduction}

\input{introduction}

\section{Background}
\label{sec:background}
\input{background}

\section{\glspl{DGMRF} on Graphs}
\label{sec:method}
\input{method}

\section{Experiments}
\label{sec:experiments}
\input{experiments}

\section{Related Work}
\label{sec:related_work}
\input{related_work}

\section{Conclusions}
\label{sec:conclusion}
\input{conclusion}

\section*{Acknowledgements}
\input{acknowledgements}

\bibliography{references}
\bibliographystyle{icml2022}

\newpage
\appendix
\onecolumn
\input{appendix}

\end{document}

%% file: macros.tex
\usepackage{amssymb} %
\usepackage{amsmath}
\usepackage{bm} %

\newcommand{\eq}[2][my_equation]{\begin{equation}\label{eq:#1}#2\end{equation}}
\newcommand{\al}[2][my_equation]{\begin{align}\label{eq:#1}#2\end{align}}
\newcommand{\als}[2][my_equation]{\begin{align}\label{eq:#1}\begin{split}#2\end{split}\end{align}}

\providecommand\refeq{} %
\renewcommand{\refeq}[1]{Eq.~\ref{eq:#1}}

\newcommand{\reffig}[1]{Figure~\ref{fig:#1}}
\newcommand{\reftab}[1]{Table~\ref{tab:#1}}
\newcommand{\refsec}[1]{section~\ref{sec:#1}}

\newcommand{\const}{\text{const.}}

\newcommand{\R}{\mathbb{R}} %

\renewcommand{\b}[1]{\bm{#1}} %

\newcommand{\indicator}[1]{\mathbb{I}_{\left\{ #1 \right\}}}
\newcommand{\norm}[2][2]{\left\lVert#2\right\rVert_{#1}} %
\DeclareMathOperator{\trace}{Tr}
\DeclareMathOperator{\diag}{diag}

\newcommand{\transpose}{\intercal}

\newcommand{\E}[2]{\mathbb{E}_{#1} \left[ #2 \right]} %
\newcommand{\normal}[2]{\mathcal{N}\left(#1, #2\right)} %
\newcommand{\normalpdf}[3]{\mathcal{N}\left(#1 \middle| #2, #3\right)} %
\newcommand{\uniform}[1]{\mathcal{U}\left(#1\right)}
\DeclareMathOperator{\var}{Var}

%% file: paper_macros.tex
\newcommand{\x}{\b{x}}
\newcommand{\z}{\b{z}}
\newcommand{\y}{\b{y}}
\newcommand{\h}{\b{h}}

\newcommand{\neigh}[1]{n(#1)}
\newcommand{\nnodes}{N}
\newcommand{\G}{\mathcal{G}}
\newcommand{\postmu}{\tilde{\b{\mu}}}
\newcommand{\postQ}{\tilde{Q}}

\newcommand{\Glayer}[1]{G^{(#1)}}
\newcommand{\glayer}[1]{g^{(#1)}}
\newcommand{\hlayer}[1]{\h^{(#1)}}
\newcommand{\blayer}[1]{\b{b}^{(#1)}}
\newcommand{\hnode}[2]{h_{#1}^{(#2)}}
\newcommand{\thetal}[1]{\theta^{(#1)}}

\newcommand{\ensemble}{($\times 10$)}

\DeclareMathOperator{\elbo}{ELBO}
\DeclareMathOperator{\sigmoid}{\text{sigmoid}}
\DeclareMathOperator{\entropy}{H}
\DeclareMathOperator{\crps}{CRPS}

\newcommand{\logdet}[1]{\log\left|\det\left(#1\right)\right|}
\newcommand{\paramq}{\left|\frac{\beta_l}{\alpha_l}\right|}

\newcommand{\ordo}[1]{\mathcal{O}\left(#1\right)}

\newcommand{\refapp}[1]{Appendix~\ref{sec:#1}}

\newcommand{\citeauth}[1]{\citeauthor{#1} (\citeyear{#1})}

\newcommand{\synthexp}{Synthetic data}
\newcommand{\wikiexp}{Wikipedia graphs}
\newcommand{\calexp}{California housing data}
\newcommand{\windexp}{Wind speed data}
\newcommand{\obsexp}{Impact of percentage of observed nodes}

%% file: acronyms.tex
\newacronym{GMRF}{GMRF}{Gaussian Markov Random Field}
\newacronym{DGMRF}{DGMRF}{Deep GMRF}

\newacronym{GNN}{GNN}{Graph Neural Network}
\newacronym{CNN}{CNN}{Convolutional Neural Network}

\newacronym{ELBO}{ELBO}{Evidence Lower Bound}
\newacronym{CG}{CG}{Conjugate Gradient}
\newacronym{MC}{MC}{Monte Carlo}

\newacronym{MAE}{MAE}{Mean Absolute Error}
\newacronym{RMSE}{RMSE}{Root Mean Squared Error}
\newacronym{CRPS}{CRPS}{Continuous Ranked Probability Score}

\newacronym{GP}{GP}{Gaussian Process}
\newacronym{LP}{LP}{Label Propagation}
\newacronym{DGP}{DGP}{Deep Graph Prior}
\newacronym{Bayes LR}{Bayes LR}{Bayesian Linear Regression}
\newacronym{IGMRF}{IGMRF}{Intrinsic GMRF}
\newacronym{LR}{LR}{Linear Regression}
\newacronym{MLP}{MLP}{Multilayer Perceptron}
\newacronym{GCN}{GCN}{Graph Convolutional Network}
\newacronym{GAT}{GAT}{Graph Attention Network}
\newacronym{SVGP}{SVGP}{Sparse Variational Gaussian Process}

%% file: cover_page.tex
\onecolumn
This paper is to appear in \textit{Proceedings of the 39th International Conference on Machine Learning}, ICML 2022.

Please cite as:
\vspace{-.9em}
\begin{Verbatim}[fontsize=\small,frame=single]
@inproceedings{graph_dgmrf,
    author = {Oskarsson, Joel and Sid{\'e}n, Per and Lindsten, Fredrik},
    booktitle = {Proceedings of the 39th International Conference on Machine Learning},
    title = {Scalable Deep {G}aussian {M}arkov Random Fields for General Graphs},
    year = {2022}
}
\end{Verbatim}

%% file: abstract.tex
Machine learning methods on graphs have proven useful in many applications due to their ability to handle generally structured data.
The framework of \glspl{GMRF} provides a principled way to define Gaussian models on graphs by utilizing their sparsity structure.
We propose a flexible \gls{GMRF} model for general graphs built on the multi-layer structure of \glsdesc{DGMRF}s, originally proposed for lattice graphs only.
By designing a new type of layer we enable the model to scale to large graphs.
The layer is constructed to allow for efficient training using variational inference and existing software frameworks for \glsdesc{GNN}s.
For a Gaussian likelihood, close to exact Bayesian inference is available for the latent field.
This allows for making predictions with accompanying uncertainty estimates.
The usefulness of the proposed model is verified by experiments on a number of synthetic and real world datasets, where it compares favorably to other both Bayesian and deep learning methods.

%% file: introduction.tex
Graphs are immensely important constructs in scientific modeling.
They show up as natural representations of technological networks, such as computer and electricity networks, but also of biological and social networks \cite{graphsandspectra, graph_signal_processing}.
As massive amounts of data are collected about these networks, the area of machine learning on graphs becomes increasingly relevant.
In this field we find probabilistic graphical models, that allow for drawing statistically sound inferences about the quantities in the graph \cite{prob_graphical_models}.
Another, more recent, family of graph-based models are \mbox{\glspl{GNN}} \cite{gnn_survey, gcn}.
These bring the flexibility and scalability of deep learning to graph-structured data.
Unifying statistically sound methods with deep learning is an important goal in the field of graph-based methods and in modern machine learning as a whole.

\begin{figure}[tb]
\centering
\includegraphics[width=\columnwidth]{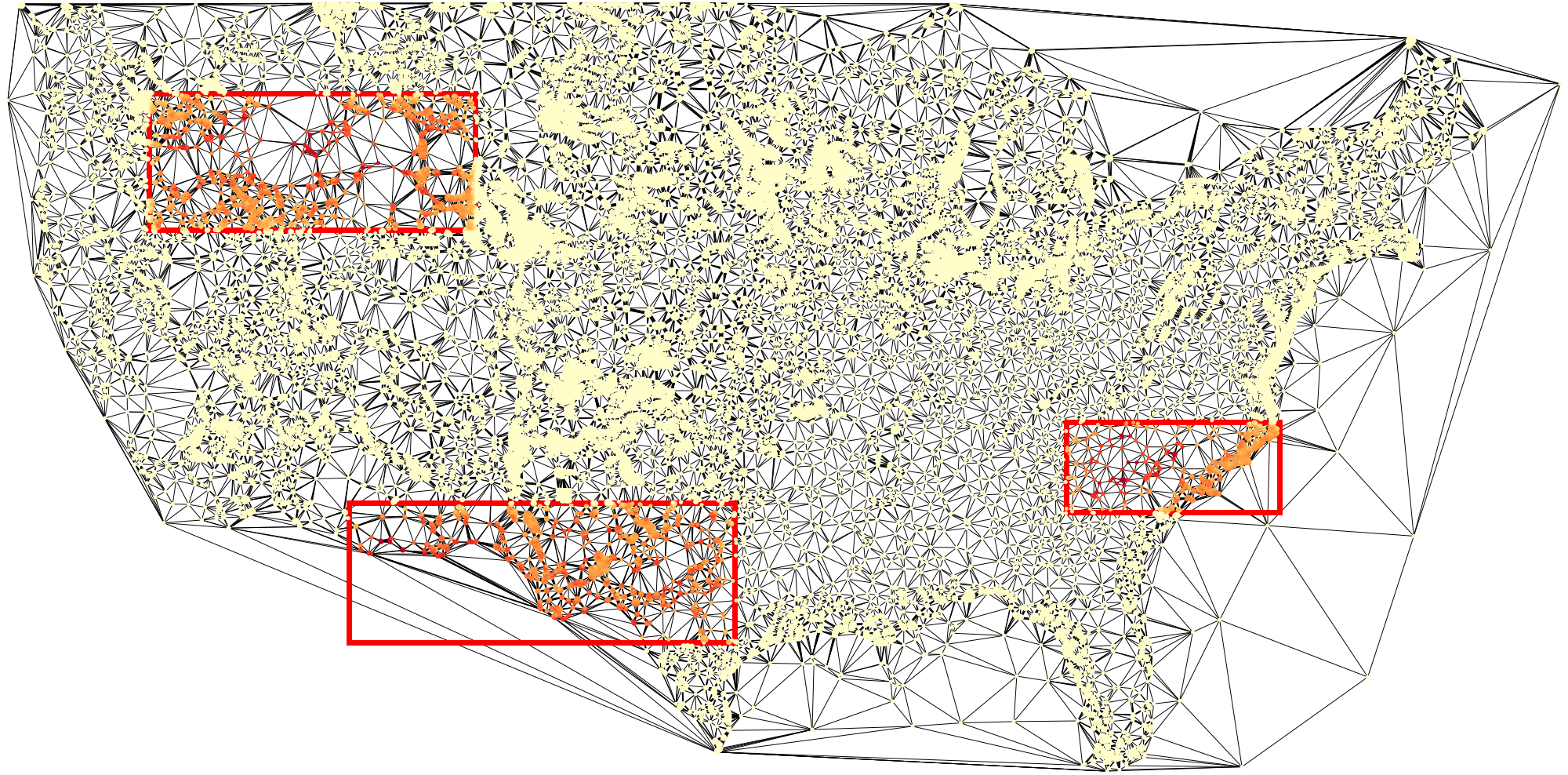}
\caption{Predictive uncertainties over a large graph. 
Nodes inside the rectangles are unobserved, resulting in higher  uncertainty (more red).
Values are marginal standard deviations, taken from the experiment with wind speed data in \refsec{wind_experiment}.}
\label{fig:example_uncertainty}
\end{figure}

There is a need for methods with a strong statistical basis, but with the scalability of \glspl{GNN}.
While some graphs, such as those describing molecules, are comparatively small, massive graphs can emerge for example from social and traffic networks.
We need efficient methods that scale also to these.
In many situations it is additionally desirable not just to produce accurate predictions, but also some measure of the uncertainty about the predictions.
An example of this is shown in \reffig{example_uncertainty}.
We are here concerned with the node-wise regression setting, where a single graph is considered and nodes are associated with real-valued targets.
We assume the full graph structure to be known. 
The main application of interest is prediction for a subset of unobserved nodes.
Such problems can for example arise when information is missing about some individuals in a social network or due to partial outages in technological networks.

Bayesian methods provide a principled way to obtain predictive uncertainty estimates, that properly account for the uncertainty in latent variables.
One type of Bayesian model that utilizes the sparsity of graphs are \acrfullpl{GMRF} \cite{gmrf_book}.
The \gls{DGMRF} framework of \citeauth{dgmrf} combines \glspl{GMRF} with \glspl{CNN} for the special case of image-structured data.
\glspl{DGMRF} can be trained efficiently and keep all useful properties of \glspl{GMRF}, such as exact Bayesian inference for the latent field.
In this paper we extend and generalize the \gls{DGMRF} framework to the general graph setting.
This requires us to design a new layer construct for \glspl{DGMRF} based on local operations over node neighborhoods.
Without making any assumptions on the graph structure we propose methods that allow the model training to scale to massive graphs.
 
Our main contributions are:
\begin{inparaenum}[1)]
\item We extend the \gls{DGMRF} framework to general graphs by designing a new type of layer construct based on \glspl{GNN}.
\item We adapt the \gls{DGMRF} training to this new setting, making use of an improved variational distribution.
\item We propose scalable methods for performing the log-determinant computations central to the training. %
\item We demonstrate properties of the resulting model on synthetic data.
\item We experiment on multiple real-world datasets, for which our model outperforms existing methods. 

\end{inparaenum}

%% file: background.tex
\subsection{\glspl{GMRF}}
\label{sec:background_gmrf}

In graphical models a set of random variables are associated with the nodes of a graph \cite{prob_graphical_models, prml}.
\glspl{GMRF} are undirected graphical models where the nodes jointly follow a Gaussian distribution.
More specifically, let $\G$ be an undirected graph with $\nnodes$ nodes concatenated in the random vector $\x \in \R^\nnodes$.
We say that $\x \sim \normal{\b{\mu}}{Q^{-1}}$ is a \gls{GMRF} with mean $\b{\mu}$ and precision matrix $Q$ w.r.t.\ the graph $\G$ iff
${Q_{i,j} \neq 0} \Leftrightarrow {j \in \neigh{i}}, \, {\forall i \neq j}$,
where $\neigh{i}$ is the exclusive neighborhood of node $i$ ($i \notin \neigh{i}$).
A \gls{GMRF} is thus a multivariate Gaussian with a precision matrix as sparse as the graph.
Note however that the covariance matrix can still be fully dense, enabling dependencies between all nodes in the graph.

Consider now the common situation where we observe ${\y = \x + \b{\epsilon}}$, $\b{\epsilon} \sim \normal{\b{0}}{\sigma^2 I}$.
A \gls{GMRF} prior on $\x$ is conjugate to this Gaussian likelihood. %
We will mainly consider the application of \glspl{GMRF} to problems where $\y$ is observed only for some nodes.
 Let $\b{m} \in \{0,1\}^\nnodes$ be a mask vector with ones in positions corresponding to the observed nodes, $\y_m = \y \odot \b{m}$ and $I_m = \diag(\b{m})$.
 The posterior for $\x$ in this setting is then given by \hbox{$\x | \y_m \sim \normal{\postmu}{\postQ^{-1}}$}, where
\begin{subequations}
\label{eq:gmrf_posterior}
\al[postx_def]{
    \postQ &= Q + \frac{1}{\sigma^{2}} I_m\\
    \postmu &= \postQ^{-1}\left(Q \b{\mu} + \frac{1}{\sigma^{2}} \y_m \right).
}
\end{subequations}
While the posterior is analytically tractable, and again a \gls{GMRF}, computing the involved entities explicitly can be a significant computational challenge for large $\nnodes$.

\subsection{\glspl{DGMRF}}
\citeauth{dgmrf} note that for an affine map ${g{:}~\R^\nnodes \rightarrow \R^\nnodes}$ a \gls{GMRF} $\x$ can be defined by 
\eq[dgmrf_def]{
    \z = g(\x) = G \x + \b{b}, \quad \z \sim \normal{\b{0}}{I}
}
where $G \in \R^{\nnodes \times \nnodes}$ is a matrix and $\b{b}$ some offset vector.
This results in a \gls{GMRF} with mean $\b{\mu} = - G^{-1}\b{b}$ and precision matrix $Q = G^\transpose G$.
Note how the direction of the mapping in \refeq{dgmrf_def} makes $\x$ implicitly defined, a different setup from other generative models mapping Gaussian noise to data.
The affine map $g$ can in turn be defined as a combination of $L$ simpler layers as $g = \glayer{L} \circ \glayer{L-1} \circ \cdots \circ \glayer{1}$, adding the depth to the Deep \gls{GMRF}.
The value of considering the layers separately is that they can be implemented implicitly, using some operation that is known to be affine.
Thus, multiple such operations can be chained without performing the expensive matrix multiplications to create $G$.

\citeauth{dgmrf} consider the special case where the entries of $\x$ are associated with pixels in an image.
They then define each $\glayer{l}$ as a 2-dimensional convolution with a filter containing trainable parameters.
Such a \gls{DGMRF} is a \gls{GMRF} w.r.t.\ a lattice graph \cite{gmrf_book}, a graph where each pixel is connected to neighboring pixels within a window determined by the filter size.
The resulting model shares much of its structure with \glspl{CNN}. 
This allows for utilizing existing deep learning frameworks for efficient convolution computations, automatic differentiation, and GPU support.

After observing some data $\y_m$ inference for the latent field $\x$ follows from \refeq{gmrf_posterior}.
To avoid inverting $\postQ$, the posterior mean $\postmu$ can be computed using the \gls{CG} method \cite{conjugate_gradient, cg_without_pain}.
Often, also the posterior marginal variances $\var\left[x_i | \y_m \right]$ are of interest.
To avoid computing the covariance matrix explicitly, an alternative is to use a \glsdesc{MC} estimate based on a set of samples from the posterior.
It is possible to use the \gls{CG} method also for efficiently drawing posterior samples \cite{gaussian_sampling}.

%% file: method.tex
\begin{figure*}[tb]
\centering
\includegraphics[width=2\columnwidth]{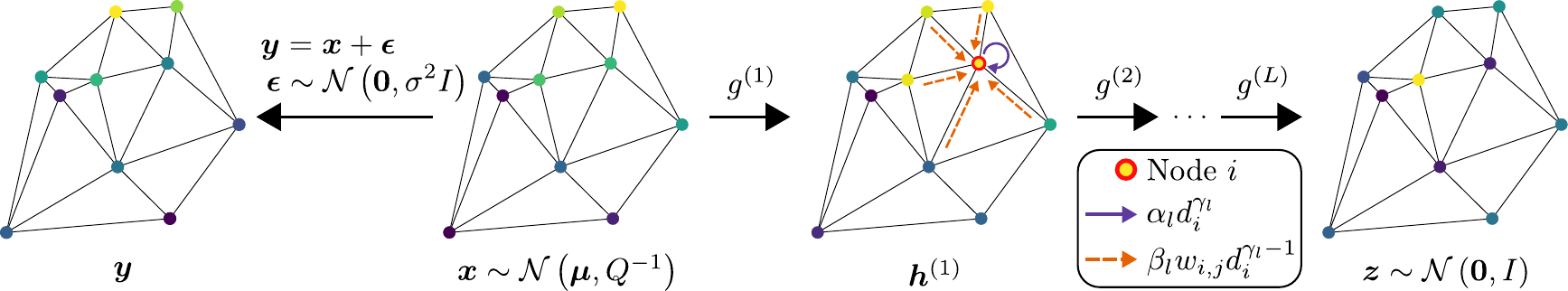}
\caption{
Overview of the graph \gls{DGMRF} model. 
The latent field $\x$ is transformed to $\z$ through $L$ affine maps $\glayer{1}, \dots, \glayer{L}$.
The data $\y$ is a noisy observation of $\x$.
In $\hlayer{1}$ we illustrate \refeq{node_update} for a single node $i$.
The node itself is weighted with $\alpha_l d_i^{\gamma_l}$ (solid purple arrow).
Each node $j$ in the neighborhood is weighted with $\beta_l w_{i,j} d_i^{\gamma_l - 1}$ (dashed orange arrows).
The value at node $i$ after the layer is the sum of all these contributions plus the bias term $b_l$.
}
\label{fig:layer_overview}
\end{figure*}

We extend the \gls{DGMRF} framework of \citeauth{dgmrf} to general graphs, removing the assumption of a lattice graph to match the general definition of a \gls{GMRF}.
To achieve this we design a new type of layer $\glayer{l}$ without any assumptions on the graph structure.
We then propose a way to train this new type of \gls{DGMRF} using scalable log-determinant computations and a new, more flexible, variational distribution.
An overview of our model is shown in \reffig{layer_overview}.

Let $\G$ be an undirected connected graph with $\nnodes$ nodes and adjacency matrix $A \in \R^{\nnodes \times \nnodes}$.
In general we consider weighted graphs with $A_{i,j} = w_{i,j} \indicator{j \in \neigh{i}}$, where $w_{i,j} = w_{j,i} > 0$ is the weight of the edge between nodes $i$ and $j$.
For unweighted graphs $w_{i,j} = 1 \, \forall i,j$.
We denote the degree of node $i$ as $d_i = \sum_j A_{i,j}$ ($= \left|\neigh{i}\right|$ in the unweighted case) and arrange all node degrees in the degree matrix $D = \diag([d_1, d_2, \dots, d_N]^\transpose)$.

\subsection{Graph layer}
\label{sec:graph_layer}
Generalizing the \gls{CNN}-based \gls{DGMRF} layers of \citeauth{dgmrf} to general graphs requires some special considerations.
It is integral to take into account the varying node degrees in the graph.
To achieve this we look to the \gls{GNN} framework and consider a linear version of a message passing neural network \cite{mpnn}.
Let ${\hlayer{l} \in \R^\nnodes}$ be the node values after layer $l$, with $\hlayer{0} = \x$ and $\hlayer{L} = \b{z}$.
An intuitive way to define the operation of $\glayer{l}$  would be to sum over the node neighborhood and weight the center node by its degree,
\eq[node_update_sum]{
    \hnode{i}{l} = b_l + \alpha_l d_i \hnode{i}{l-1} + \beta_l \smashoperator{\sum_{j \in \neigh{i}}} \hnode{j}{l-1},
}
where $\alpha_l, \beta_l$ and $b_l$ are layer-specific trainable parameters.
This operation can be viewed as a parametrized version of the graph Laplacian \cite{graphsandspectra}, a central construct in graph-based machine learning \cite{ml_on_graphs, gcn}.
As a special case of \refeq{node_update_sum} we also find the commonly used \gls{IGMRF} model \cite{gmrf_book}.
A limitation of \refeq{node_update_sum} is however that there are no parameter values that reduce the layer to an identity mapping. 
If the model would consist of a single layer, this would not have been an issue.
However, when stacking multiple layers in a deep architecture this inability introduces undesirable restrictions, in the sense that the range of attainable models is not strictly increasing as we add more layers.
To avoid this shortcoming we also consider an alternative way to define $\glayer{l}$ by instead taking the mean over the neighborhood,
\eq[node_update_mean]{
    \hnode{i}{l} = b_l + \alpha_l \hnode{i}{l-1} + \beta_l \frac{1}{d_i} \smashoperator[r]{\sum_{j \in \neigh{i}}} \hnode{j}{l-1}.
}
Unlike \refeq{node_update_sum}, this operation includes the identity mapping as a special case.

Finally, we propose a layer structure that generalizes these two ideas, making it possible for the model to learn which is the better choice for the data at hand.
Our proposed layer is defined by
\eq[node_update]{
    \hnode{i}{l} = b_l + \alpha_l d_i^{\gamma_l} \hnode{i}{l-1} + \beta_l d_i^{\gamma_l - 1} \smashoperator{\sum_{j \in \neigh{i}}} w_{i,j}\hnode{j}{l-1},
}
where we introduce another parameter $\gamma_l \in \mathopen]0,1\mathclose[$ and the optional edge weights $w_{i,j}$.
\refeq{node_update_sum} and \ref{eq:node_update_mean} are special cases as $\gamma_l$ tends to its limits.
We empirically verify the usefulness of this layer construct in \refapp{gamma_exp}.
Note also that when $\alpha_l=1$, $\beta_l=0$, $b_l=0$ and $\gamma_l \rightarrow 0$ \refeq{node_update} reduces to an identity mapping.
Another motivation for this specific layer construct is that it will enable scalable methods for computing the log-determinant of $G$, as will be explained in \refsec{log_det}.
We additionaly reparametrize the model so that $\alpha_l > 0$ and $|\beta_l| < |\alpha_l|$.
This avoids degenerate, indefinite solutions (see \refapp{reparametrization} for details) and will enable our most scalable method for log-determinant computations.

If we consider the entire vector $\hlayer{l}$, the layer corresponds to $\hlayer{l} = \glayer{l}\left(\hlayer{l-1}\right) = \Glayer{l} \hlayer{l-1} + \blayer{l}$ with
\eq[layerG]{
    \Glayer{l} = \alpha_l D^{\gamma_l} + \beta_l D^{\gamma_l - 1}A
}
and $\blayer{l} = b_l \b{1}$, where $\b{1} \in \R^\nnodes$ is a vector with all ones.
As the operation $\glayer{l}$ corresponds to a layer of a \gls{GNN} we can rely on existing software libraries for the model implementation.
Such libraries come with a number of useful properties, most importantly automatic differentiation and GPU-acceleration \cite{pytorch_geometric, spektral}.

\subsection{Variational training}
The parameters of a \gls{DGMRF} can be trained by maximizing the log marginal likelihood $\log p(\y_m|\theta)$.
This is however often infeasible as it requires computing the determinant of the posterior precision matrix $\postQ$ \cite{dgmrf}.
For large $N$ one can instead resort to variational inference, maximizing the \gls{ELBO},
\als[elbo]{
    \elbo(\theta, \phi) &= \E{q(\b{x}|\phi)}{\log p(\b{y}_m, \b{x}|\theta)} + \entropy[q(\b{x}|\phi)]\\ 
    &\leq \log p(\b{y}_m|\theta)
}
where $q$ is a variational distribution with parameters $\phi$ and $\entropy[\cdot]$ refers to differential entropy.
For a \gls{DGMRF} with a Gaussian likelihood the first term of the \gls{ELBO} is
\al[joint_likelihood]{
    \begin{split}
    &\E{q(\b{x}|\phi)}{\log p(\b{y}_m, \b{x}|\theta)} =\\ 
    &- \frac{1}{2}\E{q(\b{x}|\bm{\phi})}{g(\b{x})^\transpose g(\b{x}) + \frac{1}{\sigma^2} (\b{y}_m - \b{x})^\transpose I_m (\b{y}_m - \b{x})}\\
    &+ \logdet{G}
    - M \log \sigma
    + \const 
    \end{split}
}
where $M = \sum_{i=1}^{N} m_i$ is the number of observed nodes.
The expectation in \refeq{joint_likelihood} can be estimated using a set of samples drawn from $q$.
As $G = \Glayer{L}\Glayer{L-1}\dots\Glayer{1}$, the log-(absolute)-determinant is given by
\eq[layered_det]{
    \logdet{G} = \sum_{l=1}^{L} \logdet{\Glayer{l}}.
}
Computing this efficiently is one of the major challenges with the general graph setting, as will be discussed further in \refsec{log_det}.

The full set of model parameters $\theta$ are the trainable parameters of each layer and the noise standard deviation $\sigma$.
Maximizing the ELBO w.r.t.\ $\theta$ and $\phi$ can be done using gradient-based stochastic optimization.

\subsubsection{Variational Distribution}
A natural and useful way to choose the variational distribution is as another Gaussian $q(\x | \phi) = \normalpdf{\x}{\b{\nu}}{S S^\transpose}$.
This corresponds to defining $q$ by another affine transformation in the opposite direction of the DGMRF,
\eq[q_def]{
    \x = S \b{r} + \b{\nu}, \quad \b{r} \sim \normal{\b{0}}{I}.
}
Note the difference to \refeq{dgmrf_def} as we here parametrize the covariance matrix instead of the precision matrix.
This parametrization additionally allows for computing gradients through the sampling process, by the use of the reparametrization trick \cite{vae}.

\citeauth{dgmrf} use a simple mean field approximation with a diagonal $S$, making all components of $\x$ independent \cite{prml}.
However, we propose a more flexible $q$ by choosing
\eq[vi_layers]{
    S = \diag(\b{\xi}) \Tilde{G} \diag(\b{\tau})
}
where $\b{\xi}, \b{\tau} \in \R^\nnodes$ are vectors containing positive parameters and $\tilde{G}$ is defined in the same way as the DGMRF layer in \refeq{layerG}.
Including the matrix $\tilde{G}$ in $S$ introduces off-diagonal elements in the covariance matrix of $q$, alleviating the independence assumption between nodes.
Multiple such layers can also be used, introducing longer dependencies between nodes in the graph.
The full set of variational parameters $\phi$ is then $\b{\nu}, \b{\xi}, \b{\tau}$ and all trainable parameters from the layer(s) $\tilde{G}$.
In \refapp{vi_exp} we empirically show that \glspl{DGMRF} trained using our more flexible variational distribution consistently outperforms those trained using the simple mean field approximation.

With this choice of $S$ the entropy term of the ELBO is
\al[var_entropy]{
\begin{split}
    \entropy[&q(\b{x}|\phi)] = \logdet{S} + \const \\
    &= \logdet{\tilde{G}} + \sum_{i=1}^{\nnodes} \log \xi_i + \log \tau_i + \const
\end{split}
}
Re-using the DGMRF layer construct in $q$ has the added benefit that the techniques we develop for the log-determinant computation readily extend also to computing $\entropy[q(\b{x}|\phi)]$.

\subsection{Computing the log-determinant}
\label{sec:log_det}
Computing the necessary log-determinants in \refeq{layered_det} and \ref{eq:var_entropy} efficiently is a major challenge with the general graph setting.
The \gls{CNN}-based \gls{DGMRF} was defined on a lattice graph, which creates a special structure in $G$ and allows for finding efficient closed-form expressions for the log-determinants \cite{dgmrf}.
As we do not make any such assumptions on the graph structure we here propose new scalable methods to compute the log-determinants.

\subsubsection{Eigenvalue method}
\label{sec:eigenvalue_method}
One way to compute the log-determinant is based on the eigenvalues of the matrix.
As the determinant is given by the product of all eigenvalues,
\als[log_det_eigen]{
    \logdet{\Glayer{l}} =& \logdet{D^{\gamma_l}}\\
        &+ \logdet{\alpha_l I + \beta_l D^{- 1}A}\\
        =&\sum_{i=1}^\nnodes \gamma_l \log(d_i) + \log|\lambda_i|
} 
where $\{\lambda_i\}_{i=1}^\nnodes$ are the eigenvalues of $\alpha_l I + \beta_l D^{- 1}A$.
It can be shown\footnote{
    For an eigenvector $\b{v}_i$ of $D^{-1}A$ with eigenvalue $\lambda'_i$,
    $
    D^{-1}A\b{v}_i = \lambda'_i \b{v}_i 
    \Rightarrow  
    {(\alpha_l I + \beta_l D^{- 1}A)\b{v}_i} 
    = {(\alpha_l + \beta_l \lambda'_i) \b{v}_i}.
    $
} 
that $\lambda_i = \alpha_l + \beta_l \lambda'_i$ with $\lambda'_i$ being the $i$:th eigenvalue of the matrix $D^{-1}A$.
Since $D^{-1}A$ only depends on the graph structure, it does not change during training.
This means that we can compute the eigenvalues $\{\lambda'_i\}_{i=1}^\nnodes$ as a pre-processing step and store these for computing \refeq{log_det_eigen} during training.
Note that this is enabled by the specific layer construct that we propose.
By choosing our layer as described in \refsec{graph_layer} we manage to shift the main computational burden of log-determinant computations from the iterative training to a pre-processing step that only has to be performed once.

This method is feasible for moderate $N$, but computing all eigenvalues in pre-processing can be unreasonably slow for very large graphs.
To guarantee that the training scales to massive graphs we propose an alternative method for the log-determinant computation.

\subsubsection{Power series method}
\label{sec:power_series}
Another way to rewrite the log-determinant is
\als[log_det_reformulation]{
    \logdet{\Glayer{l}} = &N\log\left(\alpha_l\right) + \sum_{i=1}^\nnodes \gamma_l \log(d_i)\\
    &+ \logdet{I + \frac{\beta_l}{\alpha_l} \tilde{A}}
}
where we define $\tilde{A} = D^{-\frac{1}{2}} A D^{-\frac{1}{2}}$.
For computing the last term in \refeq{log_det_reformulation} we follow the approach of \citeauth{iresnet}, where an approximation of the log-determinant is constructed based on a power series,
\eq[power_series]{
    \logdet{I + \frac{\beta_l}{\alpha_l} \tilde{A}} = 
    \sum_{k=1}^\infty - \frac{1}{k} \left(-\frac{\beta_l}{\alpha_l}\right)^k \trace\left(\tilde{A}^k \right).
}
During training of the model we truncate the series at some large value $k = K$, resulting in an approximation.
We note that $\trace\left(\tilde{A}^k \right)$ only depends on the graph structure, not the model parameters.
This means that the traces for ${k \in \{1, \dots, K\}}$ can be computed as a pre-processing step.
In \refapp{power_series_details} we give further details on this pre-processing step, show that the series converges and bound the truncation error.

\subsection{Scalability}
The graph \gls{DGMRF} is defined using the graph $\G$, but it can be noted that the model is in fact not a \gls{GMRF} w.r.t.\ this graph.
Instead, an $L$-layer \gls{DGMRF} is a \gls{GMRF} w.r.t.\ the $2L$-hop graph $\G^{2L}$, which is defined by connecting all nodes that are at a distance $\leq 2L$ from each other in $\G$ (ignoring edge weights).
Equivalently, the resulting precision matrix $Q$ shares its zero-pattern with $(A + I)^{2L}$.\footnote{
Note that $\Glayer{l}$ has zeros in the same positions as $I + A$.
Since $Q = {\Glayer{1}}^\transpose \dots {\Glayer{L}} ^\transpose \Glayer{L} \dots \Glayer{1}$ it has the same zero-pattern as $(I + A)^{2L} = \sum_{k=0}^{2L} \binom{2L}{k} A^k$.
The value of $[A^k]_{i,j}$ is the number of graph walks between nodes $i$ and $j$ of length $k$.
So if $i$ and $j$ are at a distance $k' \leq 2L$ in $\G$, then $[A^{k'}]_{i,j} \neq 0$ and $Q_{i,j} \neq 0$. 
This implies that $Q$ defines a GMRF w.r.t.\ the $2L$-hop graph.
}

We consider now how the model training scales w.r.t.\ the number of layers $L$ and the graph structure.
With the proposed methods for the log-determinants these can be computed in $\ordo{\nnodes L}$.
The computational complexity is instead dominated by the application of $G$, as this operation scales with the number of edges in the graph.
To simplify the discussion, assume $d_i = d \ \forall i$, resulting in a graph with $\ordo{\nnodes d}$ edges.
Applying $L$ \gls{DGMRF} layers to this graph is then $\ordo{\nnodes d L}$.
This can be compared to explicitly creating $\G^{2L}$ in a pre-processing step, resulting in a graph with $\ordo{\nnodes d^{2L}}$ edges.
The layer structure is thus central to the scalability of the model.
Using multiple layers additionally adds flexibility in the form of more parameters and longer node dependencies in the mapping $g$.

%% file: experiments.tex
The graph \gls{DGMRF} was implemented\footnote{Our code is available at \url{https://github.com/joeloskarsson/graph-dgmrf}.} using PyTorch and the PyTorch Geometric \gls{GNN} library \cite{pytorch_geometric}. 
We evaluate the proposed model on a number of synthetic and real world datasets.
All \gls{DGMRF} training is repeated for 5 different random seeds and the \gls{DGMRF} results reported as the mean across these.
Details on the datasets and setup of each experiment are described in \refapp{exp_details}.

We consider graphs where only a fraction of the nodes are observed. 
The model parameters are trained by maximizing the ELBO using the observed nodes.
We generally treat 50\% of nodes as unobserved, chosen uniformly at random.
Unless stated otherwise we use the eigenvalue method for the log-determinant computations, as this is feasible for the majority of the considered graphs.
While the power series method could also be used in these cases, this would incur unnecessary approximations.
The trained model is then used to perform inference according to \refeq{gmrf_posterior}, utilizing the \gls{CG} method.
We do not compute the full posterior covariance matrix, but instead draw 100 posterior samples to estimate the marginal variance of each node.
Based on the posterior mean and marginal variances different evaluation metrics can then be computed for the unobserved nodes.
Using a consumer-grade GPU the full pre-processing, training, and inference procedure for a single \gls{DGMRF} model takes around 10 minutes for smaller graphs and up to one hour for the largest and most dense graphs.

\subsection{\synthexp}
\label{sec:synth_exp}
We start by considering synthetic data sampled from a model in the same class as our \gls{DGMRF}.
Based on a random graph with 3000 nodes we define \gls{DGMRF} models with 1--4 layers and some fixed, known parameters (details in \refapp{synthexp_details}).
A sample of $\x$ is then drawn from each such \gls{DGMRF} and Gaussian noise added to construct $\y$.
For this experiments we treat 25\% of the nodes as unobserved.

\begin{figure}[tb]
\centering
\includegraphics[width=\columnwidth,trim=0 2.8 0 0, clip]{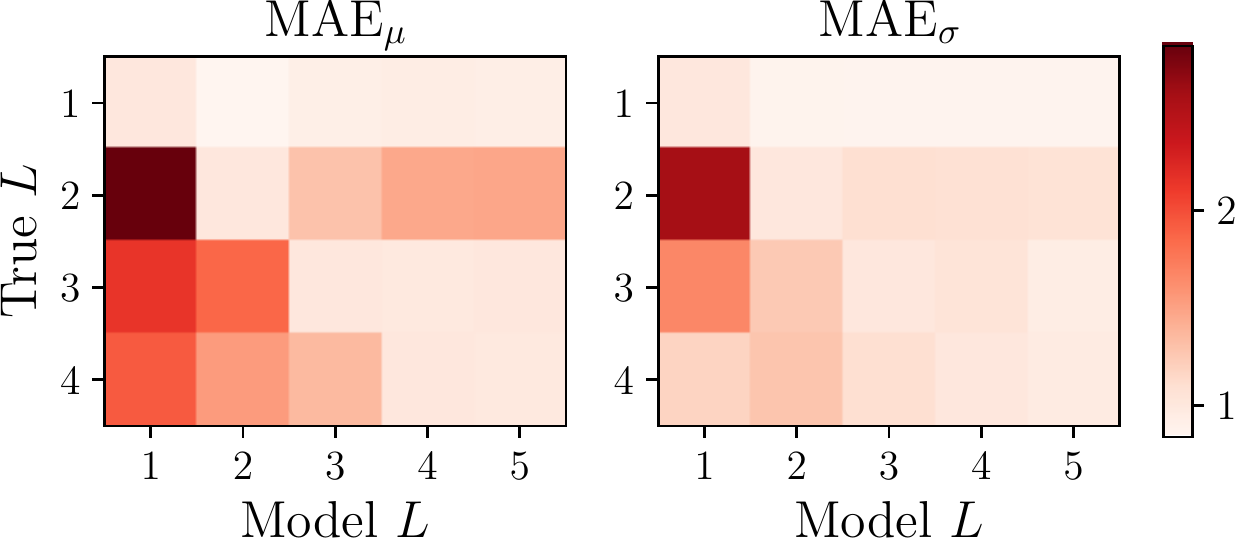}
\caption{\gls{MAE} between posterior means (left) and between posterior standard deviations (right).
The difference is computed between the true posterior of $\x$ and the posterior of each trained model.
Rows represents synthetic data sampled from DGMRFs with different number of layers.
Columns represent DGMRFs with different number of layers, trained on the synthetic data.
For easy comparison each row has been divided by the error of the model with $L$ matching the true \gls{DGMRF}, which makes the diagonal equal to 1.}
\label{fig:layer_results_mean}
\end{figure}

Using the synthetic data we train 1--5 layer \gls{DGMRF} models.
As the graph is reasonably small and we know the exact parameters of the true model we can here compute the true posterior.
We evaluate the trained models by comparing their posteriors to the known true posterior for the unobserved nodes.
The metrics used are \gls{MAE} between the posterior means and \gls{MAE} between the posterior marginal standard deviations.
Results are presented in \reffig{layer_results_mean}.

The pattern in \reffig{layer_results_mean} illustrates how a model specified with too few layers fails to learn the true posterior, as represented by the high error in the bottom left of the figure.
Without enough layers the precision matrix becomes too sparse, resulting in the model not capturing complex dependencies between nodes in the graph.
On the other hand we note how the error generally does not increase much when unnecessarily many layers are used in the model.
This indicates that deeper architectures do not impose unwanted restrictions on the model class.

\begin{figure}[tb]
\centering
\includegraphics[width=0.9\columnwidth]{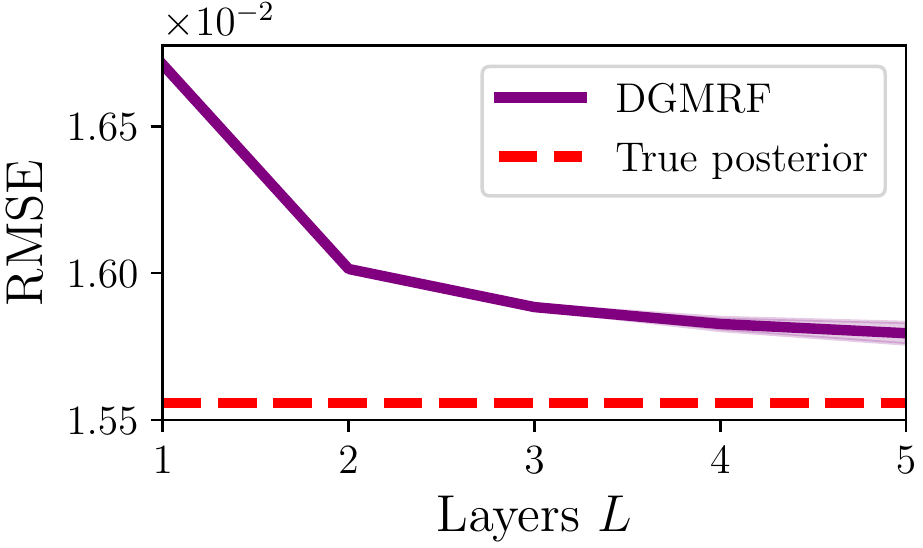}
\caption{RMSE of DGMRF models with 1--5 layers evaluated on the Mix synthetic dataset. The small shaded area corresponds to $95\%$ confidence interval across 5 random seeds. The RMSE for the true posterior mean is also shown.}
\label{fig:synth_results_lines}
\end{figure}

We consider an additional synthetic graph datasets sampled from a \gls{GMRF} that lies outside the model class of our \gls{DGMRF}.
The precision matrix of this \gls{GMRF} is specified by taking a random linear combination of some other simple precision matrices (details in \refapp{synthexp_details}).
We refer to this as the Mix dataset.
Here we compare the \gls{RMSE} of the posterior mean to the value of $\y$ for the unobserved nodes.
The results for 1--5 layer \glspl{DGMRF} are shown in \reffig{synth_results_lines}.
It is clear from the plot that \glspl{DGMRF} with more layers perform better.
More layers results in a more flexible model and a more dense precision matrix, better matching the data at hand.
An additional experiment for synthetic data constructed from another \gls{GMRF} can be found in \refapp{synth_exp_extra}.

\input{tables/wiki}

\subsection{\wikiexp}
\label{sec:wiki_experiment}
Next, we conduct experiments with three graphs based on Wikipedia pages related to different animals \cite{wikipedia_datasets}.
The three graphs vary in sparsity and number of nodes.
In these graphs the nodes represent Wikipedia pages and edges mutual links between them.
The target attribute $\y$ is the log average monthly traffic of each page.

As before we evaluate model predictions by \gls{RMSE}.
To also take the predictive uncertainty into account we additionally consider the probabilistic metric \gls{CRPS} \cite{metrics}.
Let $F_i$ be the cumulative distribution function of the predictive distribution for node $i$.
The \gls{CRPS} is then defined as
\eq[crps]{
    \crps(F_i, y_i) = - \int_{-\infty}^{\infty} \left(F_i(t) - \indicator{t \geq y_i} \right)^2 \, dt 
}
which can be interpreted as the difference between $F_i$ and a unit step located at $y_i$.
The integral in \refeq{crps} has a closed form solution when the predictive distribution is Gaussian.
We use the negative \gls{CRPS}, meaning that lower values are better.
The final metric is then computed as the mean negative CRPS over the unobserved nodes.
For baseline models that do not directly yield a predictive distribution we instead create a crude uncertainty estimate by training an ensemble of 10 models.
The mean and standard deviation across the ensemble is then used for computing the metrics.
We denote such ensembles with \ensemble{} in figures and tables.

In addition to the \gls{DGMRF} we consider a number of baseline models.
We use a regression version of \gls{LP} \cite{lp, residual_corr_gnns}, a first order \gls{IGMRF} \cite{gmrf_book} and the Graph \gls{GP} model of \citeauth{matern_graph_gp}.
It would be of interest to include a \gls{GNN} baseline, but since we do not use node features this requires special consideration.
Inspired by the CNN-based Deep Image Prior model of \citeauth{dip} we use a \gls{GNN} with random noise as input.
We denote this baseline as \gls{DGP}.
Detailed descriptions of the different baselines can be found in \refapp{baseline_details}.

The results of \glspl{DGMRF} and baseline models on the Wikipedia graphs are presented in \reftab{wiki_res} (standard deviations in \refapp{wiki_exp_extra}).
In terms of \gls{RMSE} the 1-layer \gls{DGMRF} performs similar or better to the baseline models. 
Adding more layers also seems beneficial, as the best performance is reached at $L=5$.
It can be noted that the diameters of the considered graphs are $\approx 10$, meaning that a 5-layer \gls{DGMRF} corresponds to an almost fully dense precision matrix.
Note that our framework allows us to define such a model without ever working with that large matrix explicitly.
With close to exact posterior inference the \gls{DGMRF} model also gives principled uncertainty estimates, which is reflected in the favorable \gls{CRPS} values.

\input{tables/cal}
\subsection{\calexp}
\label{sec:cal_experiment}
We here experiment on the classical California housing dataset \cite{california}.
This dataset contains median house values of 20\,640 housing blocks located in California.
Based on their spatial coordinates we create a sparse graph by Delaunay triangulation \cite{triangulations_book}. %
We also weight the graph by the inverse distances between nodes, here showcasing the ability of our model to work with a weighted adjacency matrix.
The California housing dataset additionally contains socio-economic features associated with the housing blocks. 
To handle these node features in the \gls{DGMRF} we use an auxiliary linear model, similarly to \citeauth{dgmrf}.
The linear model is also given a Bayesian treatment with its own variational distribution trained jointly with the rest of the model.

For this experiment additional deep learning baselines are used.
We use a standard \gls{MLP} and two \glspl{GNN}, \gls{GCN} \cite{gcn} and \gls{GAT} \cite{gat}.
A Bayesian \gls{LR} baseline is also included.
To compare the graph-based methods to a direct spatial approach we also consider a \gls{SVGP} model \cite{svgp}.

In \reftab{cal_res} we report results with and without using the node features.
The results demonstrate how utilizing both the graph structure and node features allows our \gls{DGMRF} to accurately model the data.
Even without node features the probabilistic predictions of our \gls{DGMRF} result in the lowest \gls{CRPS}.
While the main point of this experiment is to compare purely graph-based methods, it also serves as an example of how our \gls{DGMRF} can be applied to spatial data.
It should however be noted that a graph based on spatial positions is an approximation.
If only the graph is used by the model, some spatial information is being lost.
Methods working directly with the continuous coordinates can in some cases be more suitable for this type of problem, but often come with computational challenges \cite{svgp}.

\input{tables/wind}
\subsection{\windexp}
\label{sec:wind_experiment}
To test the scalability of our model we experiment on a large dataset containing wind speeds.
The dataset contains the average wind speed for 126\,652 sites around the US, based on a state-of-the-art meteorological simulation \cite{wind_toolkit}.
Similar pre-processing and experiment setup is used as for the California housing data.

Due to the large size of the graph we here use the power series method for log-determinant computations.
Both a random and spatial observation mask were considered.
The spatial mask can be seen in \reffig{example_uncertainty}.
Results for the wind speed dataset are presented in \reftab{wind_res}.
The experiment showcases the ability of our model to scale to large graphs without sacrificing predictive performance.

\begin{figure}[htb]
\centering
\includegraphics[width=\columnwidth]{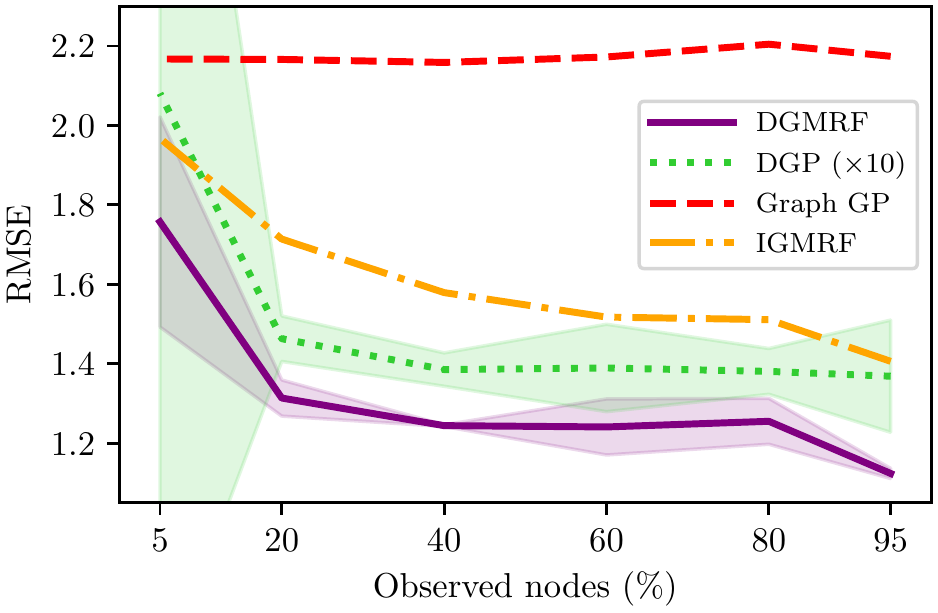}
\caption{RMSE on the Crocodile Wikipedia graph for different percentages of observed nodes. The shaded area corresponds to $95\%$ confidence interval, evaluated across different random seeds.}
\label{fig:fraction_croc_rmse}
\end{figure}

\subsection{\obsexp}
\label{sec:fraction_labeled}
In the next experiment we investigate how the model performance is impacted by the percentage of observed nodes.
We repeat the experiment on the Crocodile Wikipedia graph with 5\%--95\% of the nodes observed and a 3-layer \gls{DGMRF}.
The RMSE of the \gls{DGMRF} and a number of baseline models are presented in \reffig{fraction_croc_rmse}.
Similar plots for \gls{CRPS} and for the synthetic Mix dataset can be found in \refapp{obs_exp_extra}.
From \reffig{fraction_croc_rmse} we note how the the \gls{DGMRF} performs well at 20\% observed nodes and improves somewhat as even more are observed.
When only 5\% of the nodes are observed the resulting \gls{RMSE} is substantially higher and the model seems to become more sensitive to the random seed used.

%% file: tables/wiki.tex
\begin{table*}[tb]
\caption{Results on the three Wikipedia datasets. The best value of each metric is marked with bold and second best underlined. The last column denotes the number of trainable parameters in each model, not including variational parameters.}
\label{tab:wiki_res}
\vskip 0.15in
\centering
\begin{small}
\begin{sc}
\begin{tabular}{@{}lccccccc@{}}
\toprule
 & \multicolumn{2}{c}{\textbf{Chameleon}} & \multicolumn{2}{c}{\textbf{Squirrel}} & \multicolumn{2}{c}{\textbf{Crocodile}} & \\
\cmidrule(lr){2-3} \cmidrule(lr){4-5} \cmidrule(lr){6-7}
\textbf{Model} & RMSE & CRPS & RMSE & CRPS & RMSE & CRPS & \textbf{\# Parameters} \\ \midrule
Graph GP & 2.115 & 1.216 & 1.772 & 1.001 & 2.169 & 1.251 & 4 \\
LP & 2.102 & - & 1.930 & - & 2.014 & - & - \\
IGMRF & 1.805 & 1.030 & 1.718 & 0.986 & 1.526 & 0.939 & 2 \\ 
DGP \ensemble{} & 1.613 & 1.067 & \underline{1.400} & 0.984 & 1.308 & 0.786 & $\approx 10^4$ \\
\midrule
DGMRF, $L=1$ & 1.589 & 0.883 & 1.643 & 0.924 & 1.311 & 0.704 & 5 \\
DGMRF, $L=3$ & \underline{1.511} & \underline{0.835} & 1.493 & \underline{0.837} & \underline{1.228} & \underline{0.652} & 13 \\
DGMRF, $L=5$ & \textbf{1.465} & \textbf{0.804} & \textbf{1.374} & \textbf{0.765} & \textbf{1.169} & \textbf{0.614} & 21 \\ \bottomrule
\end{tabular}
\end{sc}
\end{small}
\vskip -0.1in
\end{table*}

%% file: tables/cal.tex
\begin{table}[tb]
\caption{Results on the California housing dataset. All metrics are listed multiplied by a factor 100. When using no features only the graph or spatial coordinates are utilized.}
\label{tab:cal_res}
\vskip 0.15in
\centering
\begin{small}
\begin{sc}
\begin{tabular}{@{}lllll@{}}
\toprule
 & \multicolumn{2}{c}{\textbf{No features}} & \multicolumn{2}{c}{\textbf{Features}} \\ 
 \cmidrule(lr){2-3} \cmidrule(lr){4-5}
\textbf{Model} & \multicolumn{1}{c}{RMSE} & \multicolumn{1}{c}{CRPS} & \multicolumn{1}{c}{RMSE} & \multicolumn{1}{c}{CRPS} \\ \midrule
Bayes LR & 13.319 & 7.521 & 8.872 & 4.834 \\
MLP \ensemble{} & 11.086 & 7.915 & 7.094 & 4.525 \\
GCN \ensemble{}& 8.760 & 5.683 & 6.837 & 4.273 \\
GAT \ensemble{}& 9.166  & 6.049 & 6.788 & 4.348 \\
SVGP & 10.172 & 5.689 & 7.287 & 3.930 \\
Graph GP & 11.202 & 6.350 & \multicolumn{2}{c}{-} \\
LP & 6.989 & \multicolumn{1}{c}{-} & \multicolumn{2}{c}{-} \\
IGMRF & 6.989 & 3.841 & \multicolumn{2}{c}{-} \\ \midrule
DGMRF, $L=1$ & \underline{6.909} & 3.665 & 5.894 & 3.078 \\
DGMRF, $L=2$ & \textbf{6.853} & \textbf{3.651} & \underline{5.810} & \underline{3.041} \\
DGMRF, $L=3$ & \textbf{6.853} & \underline{3.656} & \textbf{5.804} & \textbf{3.039} \\ \bottomrule
\end{tabular}
\end{sc}
\end{small}
\vskip -0.1in
\end{table}

%% file: tables/wind.tex
\begin{table}[tb]
\caption{Results on the wind speed dataset. 
In \textit{Spatial Mask} three rectangular areas were masked out, with nodes inside these treated as unobserved. 
For \textit{Random Mask} half of the nodes were treated as unobserved, chosen uniformly at random.}
\label{tab:wind_res}
\vskip 0.15in
\centering
\begin{small}
\begin{sc}
\begin{tabular}{@{}lllll@{}}
\toprule
 & \multicolumn{2}{c}{\textbf{Spatial Mask}} & \multicolumn{2}{c}{\textbf{Random Mask}} \\
 \cmidrule(lr){2-3} \cmidrule(lr){4-5}
\textbf{Model} & \multicolumn{1}{c}{RMSE} & \multicolumn{1}{c}{CRPS} & \multicolumn{1}{c}{RMSE} & \multicolumn{1}{c}{CRPS} \\ \midrule
Bayes LR & 1.167 & 0.677 & 0.948 & 0.524\\
MLP \ensemble{}& 1.176 & 0.815 & 0.653 & 0.396 \\
GCN \ensemble{}& 1.116 & 0.719 & 0.606 & 0.357 \\
GAT \ensemble{}& 1.082 & 0.650 & 0.622 & 0.362 \\
LP & \textbf{0.979} & \multicolumn{1}{c}{-} & \underline{0.311} & \multicolumn{1}{c}{-} \\
IGMRF & \textbf{0.980} & \textbf{0.610} & \underline{0.311} & \underline{0.160} \\ \midrule
DGMRF, $L=1$ & 1.246 & 0.700 & \textbf{0.272} & \textbf{0.123} \\
DGMRF, $L=2$ & 1.083 & 0.616 & \textbf{0.273} & \textbf{0.123} \\
DGMRF, $L=3$ & \underline{1.075} & \underline{0.612} & \textbf{0.272} & \textbf{0.123} \\ \bottomrule
\end{tabular}
\end{sc}
\end{small}
\vskip -0.1in
\end{table}

%% file: related_work.tex
Using sparse linear algebra computations, including the sparse Cholesky decomposition, \glspl{GMRF} can be scaled to large graphs \citep[Section 2.3--2.4]{gmrf_book}.
These methods do however rely on $Q$ being highly sparse, which is not always the case for multi-layer \glspl{DGMRF}.
Our \gls{GNN}-based framework gives us additional benefits in the form of automatic differentiation and GPU-acceleration.
Another approach to automatic differentiation for \glspl{GMRF} is that of \citeauth{banded_ops_autodiff}, where the sparse Cholesky operator is endowed with a method for reverse-mode differentiation.

\glspl{GMRF} have close connections to \glspl{GP} and multiple attempts have been made to also define \glspl{GP} on graphs.
\citeauth{gp_on_graph} define a version of \glspl{GP} on comparably small graphs, but as with regular \glspl{GP} scaling to large datasets is a challenge.
\citeauth{matern_graph_gp} propose a more scalable graph \gls{GP} by making use of variational inference and stochastic optimization.
Their method does however require computing eigenvalues of the graph Laplacian.
For our \gls{DGMRF} this can be circumvented by the power series method discussed in \refsec{power_series}, but this technique does not readily extend to the graph \gls{GP} case.
In our experiments this graph \gls{GP} also seems to suffer in predictive performance due to the approximations necessary to make it scale to very large graphs.
\citeauth{deep_gp_on_graph} define deep \glspl{GP} \cite{doubly_stochastic_gps} on graphs.
They do however consider a different setting, where multiple signals are available for the same graph.
\glspl{GP} on graphs can also be defined as solutions to stochastic partial differential equations.
This approach is used by \citeauth{graph_gp_via_SPDEs} to construct spatio-temporal graph \glspl{GP} and by \citeauth{matern_fields_on_graphs} to define a \gls{GP} on both nodes and edges of the graph.
Also noteworthy is that \glspl{GMRF} can be viewed as approximations to \glspl{GP} in both the euclidean and graph settings \cite{gf_gmrf_link, matern_graph_gp}.

\glspl{GNN} are more scalable graph models based on deep learning \cite{gnn_survey, gcn}.
While \glspl{GNN} have proven to give accurate predictions in many settings they lack the probabilistic interpretation of \glspl{GMRF} and \glspl{GP}.
Our model utilizes the computational benefits of \glspl{GNN} without sacrificing the rigorous statistical basis of \glspl{GMRF}.
Attempts have also been made to combine \glspl{GNN} with probabilistic approaches, for example by modeling the correlation of node residuals as jointly Gaussian \cite{residual_corr_gnns}.

One interpretation of \glspl{DGMRF} is as a linear version of a normalizing flow model \cite{dgmrf, real_nvp}.
Normalizing flows have been defined for graphs, but mainly for generating the graph structure \cite{graph_nf, graph_nvp}.
\citeauth{dgmrf} define a non-linear version of \gls{DGMRF} (a normalizing flow) in the \gls{CNN}-setting, but demonstrate no convincing results on its usefulness.
Preliminary experiments on a non-linear version of our graph \gls{DGMRF} indicate similar poor results.

%% file: conclusion.tex
In this paper we have presented \glspl{DGMRF} for general graphs.
Through the computational framework of \glspl{GNN} and scalable log-determinant computations the model can be applied to large graphs.
In experiments the proposed model compares favorably to other both Bayesian and deep learning methods.
We have only considered Gaussian likelihoods, but extensions to non-conjugate settings could be of interest (for example node classification).
Other directions for future work involve better ways to make use of node features and methods for jointly learning also the graph structure.

%% file: acknowledgements.tex
This research is financially supported by the Swedish Research Council via the project
\emph{Handling Uncertainty in Machine Learning Systems} (contract number: 2020-04122),
the Wallenberg AI, Autonomous Systems and Software Program (WASP) funded by the Knut and Alice Wallenberg Foundation,
and
the Excellence Center at Linköping--Lund in Information Technology (ELLIIT).

%% file: appendix.tex
\section{Additional Results}
\label{sec:additional_res}
\input{appendices/additional_results}

\section{Reparametrization}
\label{sec:reparametrization}
\input{appendices/reparametrization}

\section{Details on the Power Series Log-Determinant Computation}
\label{sec:power_series_details}
\input{appendices/power_series}

\section{Baseline Models}
\label{sec:baseline_details}
\input{appendices/baselines}

\section{Details on Experiments and Datasets}
\label{sec:exp_details}
\input{appendices/experiment_details}

\bibliographyapp{references}
\bibliographystyleapp{icml2022}

%% file: appendices/additional_results.tex
In this appendix we present additional experiments that verify properties of our model.
We also present some supplementary results from the different experiments in \refsec{experiments}.

\subsection{Trainable $\gamma_l$}
\label{sec:gamma_exp}
In \refsec{graph_layer} we propose a new \gls{DGMRF} layer that generalizes \refeq{node_update_sum} and \ref{eq:node_update_mean} by introducing the $\gamma_l$-parameter.
To verify the usefulness of this parametrization, we perform a number of experiments with fixed $\gamma_l=1$ (\refeq{node_update_sum}), $\gamma_l=0$ (\refeq{node_update_mean}), and with $\gamma_l$ being trainable (\refeq{node_update}, our proposed layer).
We train and evaluate \glspl{DGMRF} with 1, 3 and 5 layers on multiple datasets.
To decouple the parametrization of the \gls{DGMRF} layers from any layers $\tilde{G}$ used in our variational distribution we here only use a mean field approximation.
Apart from this we use the same experiment setups as described in \refsec{experiments} and \refapp{exp_details} for each dataset.

Results from the experiments are reported in \reftab{gamma_res}.
It is clear that \glspl{DGMRF} with a trainable $\gamma_l$-parameter generally models the data more accurately.
In practice it can also be observed that $\gamma_l$ often takes values that are not close to neither 0 nor 1, further indicating the usefulness of the added flexibility.
Comparing the models with fixed $\gamma_l$, the version with $\gamma_l = 0$ generally performs better, in particular in multi-layer cases.
We have observed that with $\gamma_l = 1$ the training easily becomes unstable. 
For the Chameleon data the training of the multi-layer models does not even converge, so these results are left out.
In \reftab{gamma_res} it is also noteworthy that the \gls{DGMRF} with $L=3$ and $\gamma_l=1$ corresponds to the true model for the Synthetic \gls{DGMRF} data.
Despite this the \gls{DGMRF} with trainable $\gamma_l$ ends up closer to the true posterior.
We believe the reason for this is the problematic training of the $\gamma_l = 1$ models, preventing the trained \gls{DGMRF} from reaching parameter values close to the true ones.
On the synthetic \gls{DGMRF} data the $\gamma_l = 1$ model especially seems to suffer from the independence assumptions in the mean field approximation.
If trained using our more flexible variational distribution the error of this model is reduced substantially\footnote{
We re-trained the 3-layer model on the synthetic \gls{DGMRF} data using our variational distribution including 2 \gls{DGMRF} layers (with trainable $\gamma_l$).
The corresponding $\text{MAE}_\mu$ is then $0.266 \pm 0.016$ for $\gamma_l = 1$, $0.395 \pm 0.000$ for $\gamma_l = 0$ and $0.149 \pm 0.003$ for $\gamma_l$ trainable.
}.

\input{tables/gamma}

\subsection{Variational distribution}
\label{sec:vi_exp}
To evaluate our improved variational distribution we conduct additional experiments on a few of our considered datasets.
We train \gls{DGMRF} models with 1, 3 and 5 layers on the synthetic 3-layer \gls{DGMRF} data, the Chameleon Wikipedia data and the California housing data without features.
We train each \gls{DGMRF} model using three different variational distributions: 
\begin{itemize}
    \item Mean Field (MF) approximation (used by \citeauth{dgmrf}),
    \item $\tilde{G}$ corresponding to 1 \gls{DGMRF} layer (ours, see \refeq{vi_layers}),
    \item $\tilde{G}$ corresponding to 2 \gls{DGMRF} layers.
\end{itemize}
We then use the standard experiment setup for each dataset.
As we have seen no use for more than 3-layer models on the California housing dataset we there exclude the 5-layer \gls{DGMRF}.

\reftab{vi_res1} and \ref{tab:vi_res2} list results for the different variational distributions.
Our more flexible distribution shows a clear advantage over the simple mean field approximation.
We also report the converged \gls{ELBO} values of each model.
The addition of \gls{DGMRF} layers also in the variational distribution results in higher \glspl{ELBO} overall.
This indicates that the learned variational distribution is closer to the true posterior in terms of Kullback--Leibler divergence.
The higher \gls{ELBO} can also be explained by a higher log marginal likelihood for the learned model parameters, which would be in line with the improved metric values.
While the use of 1 layer in the variational distribution is an improvement over the mean field approximation, there is no pronounced benefit of 2 layers.
This motivates our choice to use 1 layer in remaining experiments. 

\input{tables/vi_results}

\subsection{\synthexp}
\label{sec:synth_exp_extra}

\begin{figure}[tb]
\centering
\includegraphics[width=0.5\columnwidth]{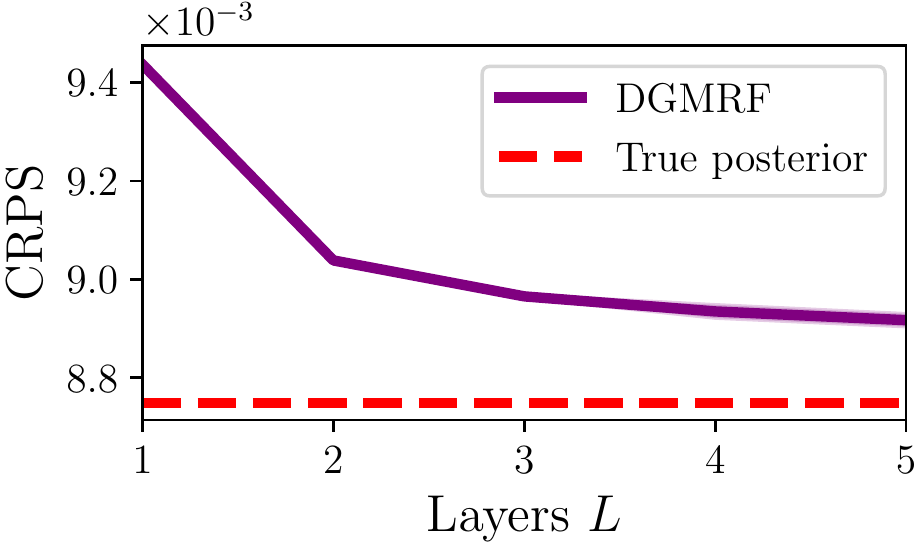}
\caption{CRPS of DGMRF models with 1--5 layers evaluated on the Mix synthetic dataset. The shaded area corresponds to $95\%$ confidence interval across 5 random seeds. The CRPS for the true posterior is also shown.}
\label{fig:synth_results_lines_crps}
\end{figure}

In \refsec{synth_exp} \gls{DGMRF} models were trained and evaluated on the synthetic Mix dataset.
We present additional results from this experiment in \reffig{synth_results_lines_crps}, where we evaluate the \gls{CRPS} of the model predictions.

\begin{figure}[tb]
\centering
\subfigure[RMSE]{\includegraphics[width=0.49\linewidth]{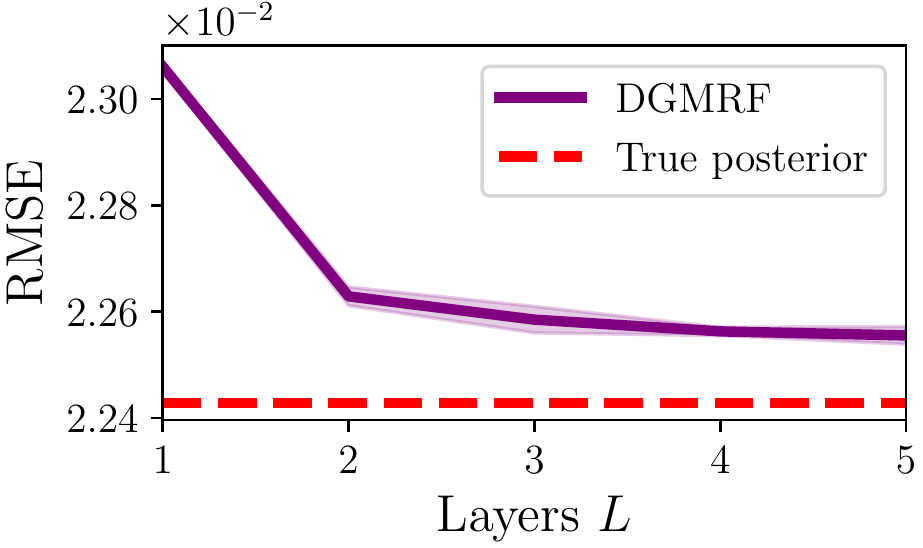}}
\subfigure[CRPS]{\includegraphics[width=0.49\linewidth]{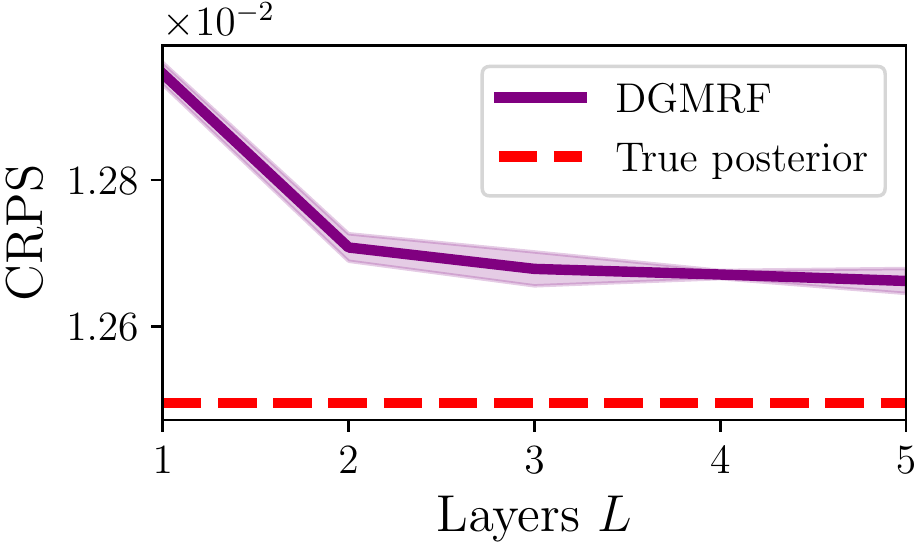}}
\caption{RMSE and CRPS of DGMRF models with 1--5 layers evaluated on the Dense synthetic datasets. The shaded area corresponds to $95\%$ confidence interval across 5 random seeds. The metric values for the true posterior is also shown.}
\label{fig:synth_results_lines_dense}
\end{figure}

We consider another synthetic dataset to evaluate how well multi-layer \glspl{DGMRF} can model data from a \gls{GMRF} with a more dense precision matrix.
Based on a random graph $\G$ we create the 3-hop graph $\G^3$.
We then draw a sample from a 1-layer DGMRF defined w.r.t. $\G^3$.
Gaussian noise is then added and 25\% of the nodes treated as unobserved.
We call this the Dense synthetic dataset.
Similar to the experiment on the Mix data we train 1--5 layer \glspl{DGMRF}.
Note that the models we train are defined on $\G$ rather than $\G^3$.
The true model for the data also here lies outside the model class of \glspl{DGMRF} defined on $\G$.
Results for the Dense dataset are presented in \reffig{synth_results_lines_dense}.
With more layers the \gls{DGMRF} can come close to the true posterior, even when defined on a more sparse graph. 
The performance of the \gls{DGMRF} models clearly improves up until 3 layers, at which point the sparsity of the precision matrix matches that of the true model.

\subsection{\wikiexp}
\label{sec:wiki_exp_extra}
For the experiments on Wikipedia graphs (\refsec{wiki_experiment}) we list the standard deviations across 5 random seeds in \reftab{wiki_std}.
The random seeds only affect the model training and the set of unobserved nodes is kept fixed.

\input{tables/wiki_std}

\subsection{California housing and wind speed data}
\label{sec:spatial_exp_extra}
For the experiments on the California housing and wind speed data (\refsec{cal_experiment} and \ref{sec:wind_experiment}) we list the standard deviations across 5 random seeds in \reftab{spatial_std}.

\input{tables/spatial_std}

\subsection{\obsexp}
\label{sec:obs_exp_extra}
\begin{figure}[tb]
\centering
\includegraphics[width=0.5\columnwidth]{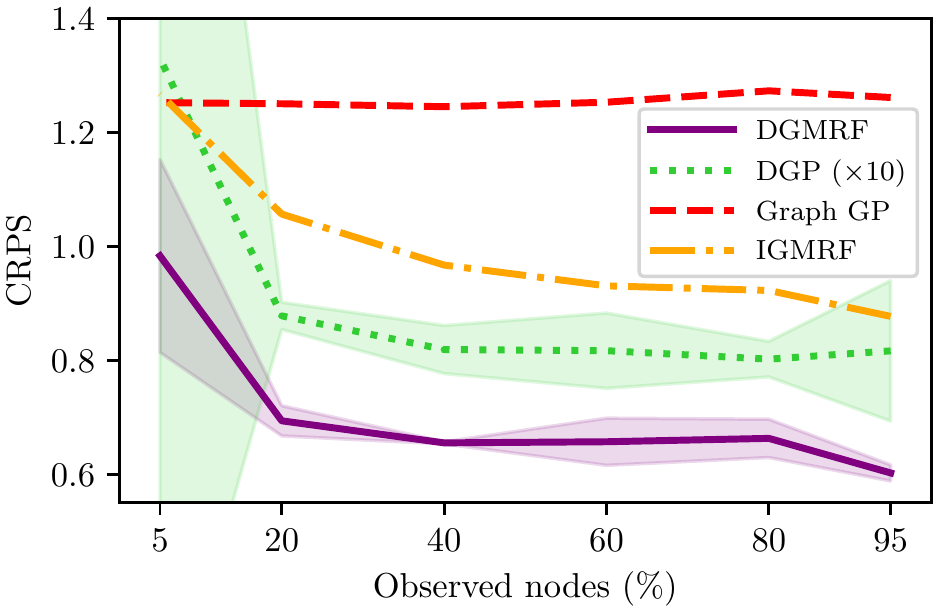}
\caption{CRPS on the Crocodile Wikipedia graph for different percentages of observed nodes. The shaded area corresponds to $95\%$ confidence interval, evaluated across different random seeds.}
\label{fig:label_fraction_croc_crps}
\end{figure}

\begin{figure}[tb]
\centering
\subfigure[RMSE]{\includegraphics[width=0.49\linewidth]{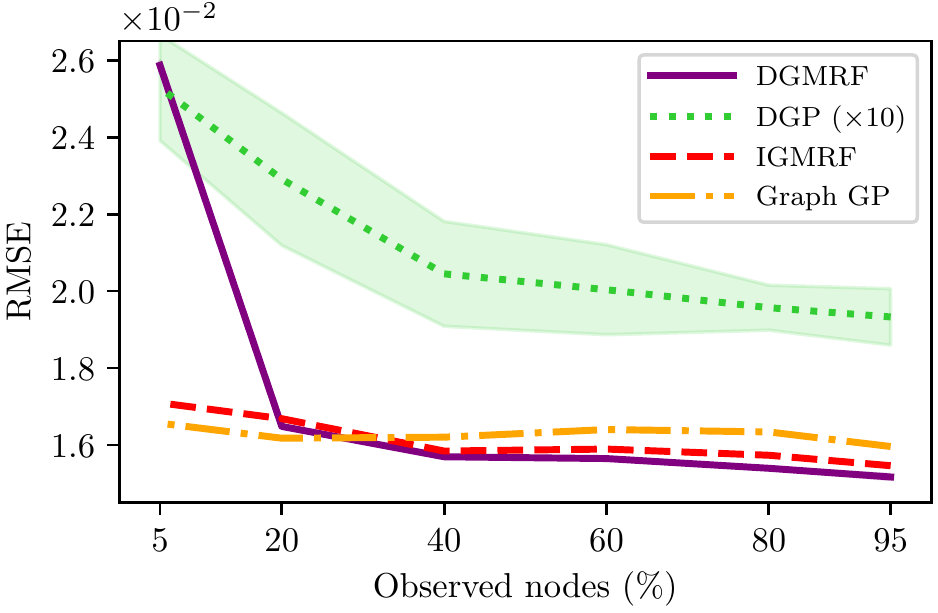}}
\subfigure[CRPS]{\includegraphics[width=0.49\linewidth]{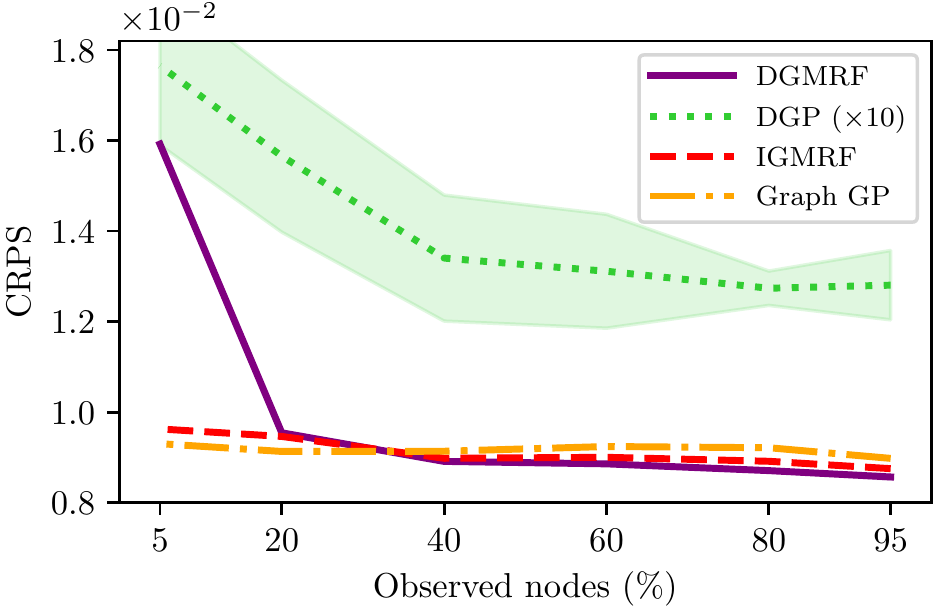}}
\caption{RMSE and CRPS on the Mix synthetic graph for different percentages of observed nodes. The shaded area corresponds to $95\%$ confidence interval, evaluated across different random seeds.}
\label{fig:label_fraction_mix}
\end{figure}

Here we present additional results for the experiment in \refsec{fraction_labeled}, where model performance for different percentages of observed nodes is investigated.
\reffig{label_fraction_croc_crps} shows for the Crocodile Wikipedia graph how the \gls{CRPS} changes with the percentage of observed nodes.
The same experiment was also repeated using the synthetic Mix dataset and the results from this can be found in \reffig{label_fraction_mix}.

%% file: tables/gamma.tex
\begin{table*}[bp]
\caption{
Results comparing a fixed and trainable $\gamma_l$-parameter.
\gls{DGMRF} models with $L \in \{1,3,5\}$ layers have been considered.
The datasets used are: synthetic data sampled from a 3-layer \gls{DGMRF} (earlier described in \refsec{synth_exp}), the synthetic Mix dataset and the Chameleon Wikipedia data.
For the first of these we compare the model to the true posterior using MAE of posterior mean and marginal standard deviations.
All metrics are reported as mean$\pm$standard deviation across 5 random seeds.
For readability some metrics are listed multiplied by a factor 100.
}
\label{tab:gamma_res}
\vskip 0.15in
\centering
\begin{small}
\begin{sc}

\begin{tabular}{@{}llcccccc@{}}
\toprule
 &  & \multicolumn{2}{c}{\textbf{Synth. DGMRF ($L=3$)}} & \multicolumn{2}{c}{\textbf{Mix}} & \multicolumn{2}{c}{\textbf{Chameleon}} \\ 
 \cmidrule(lr){3-4} \cmidrule(lr){5-6} \cmidrule(lr){7-8}
\textbf{\gls{DGMRF}} & $\bm{\gamma_l}$ & $\text{MAE}_{\mu}$ ($\times 100$) & $\text{MAE}_{\sigma}$ ($\times 100$) & RMSE ($\times 100$) & CRPS ($\times 100$) & RMSE & CRPS \\ 
\midrule
$L=1$ & $\gamma_l = 1$ & \textbf{0.335$\bm{\pm}$0.000} & \textbf{0.355$\bm{\pm}$0.004} & 1.752$\pm$0.001 & 0.988$\pm$0.000 & 1.841$\pm$0.001 & 1.065$\pm$0.001 \\ 
 & $\gamma_l = 0$ & 0.441$\pm$0.000 & 0.440$\pm$0.005 & 1.689$\pm$0.000 & 0.955$\pm$0.000 & \textbf{1.617$\bm{\pm}$0.001} & \textbf{0.899$\bm{\pm}$0.001} \\
 & trainable & \textbf{0.335$\bm{\pm}$0.000} & \textbf{0.355$\bm{\pm}$0.006} & \textbf{1.687$\bm{\pm}$0.000} & \textbf{0.954$\bm{\pm}$0.000} & \textbf{1.617$\bm{\pm}$0.001} & \textbf{0.899$\bm{\pm}$0.001} \\
 \midrule
$L=3$ & $\gamma_l = 1$ & 0.370$\pm$0.008 & 0.446$\pm$0.024 & 1.945$\pm$0.002 & 1.050$\pm$0.001 & - & - \\ 
 & $\gamma_l = 0$ & 0.404$\pm$0.000 & 0.372$\pm$0.010 & 1.587$\pm$0.000 & 0.896$\pm$0.000 & 1.634$\pm$0.001 & 0.904$\pm$0.001 \\
 & trainable & \textbf{0.163$\bm{\pm}$0.002} & \textbf{0.240$\bm{\pm}$0.010} & \textbf{1.584$\bm{\pm}$0.001} & \textbf{0.894$\bm{\pm}$0.000} & \textbf{1.527$\bm{\pm}$0.002} & \textbf{0.845$\bm{\pm}$0.001} \\
\midrule
$L=5$ & $\gamma_l = 1$ & 0.935$\pm$0.056 & 0.904$\pm$0.034 & 5.188$\pm$0.698 & 1.753$\pm$0.035 & - & - \\ 
 & $\gamma_l = 0$ & 0.414$\pm$0.000 & 0.384$\pm$0.005 & 1.581$\pm$0.000 & 0.892$\pm$0.000 & 1.584$\pm$0.001 & 0.861$\pm$0.000 \\
 & trainable & \textbf{0.184$\bm{\pm}$0.005} & \textbf{0.229$\bm{\pm}$0.002} & \textbf{1.578$\bm{\pm}$0.001} & \textbf{0.890$\bm{\pm}$0.001} & \textbf{1.489$\bm{\pm}$0.063} & \textbf{0.813$\bm{\pm}$0.039} \\
 \bottomrule
\end{tabular}

\end{sc}
\end{small}
\vskip -0.1in
\end{table*}

%% file: tables/vi_results.tex
\begin{table*}[tbp]
\caption{
Results comparing models trained using different variational distributions.
\gls{DGMRF} models with $L \in \{1,3,5\}$ layers have been trained on synthetic data sampled from a 3-layer \gls{DGMRF}.
We compare the models to the true posterior using MAE of posterior mean and marginal standard deviations.
All values are reported as mean$\pm$standard deviation across 5 random seeds.
Note that higher \gls{ELBO} values are better.
}
\label{tab:vi_res1}
\vskip 0.15in
\centering
\begin{small}
\begin{sc}

\begin{tabular}{@{}llccc@{}}
\toprule
&  & \multicolumn{3}{c}{\textbf{Synth. DGMRF ($L=3$)}} \\ 
\cmidrule(lr){3-5}
\textbf{DGMRF} & \textbf{VI Layers} & $\text{MAE}_{\mu}$ ($\times 100$) & $\text{MAE}_{\sigma}$ ($\times 100$) & ELBO \\ \midrule
$L=1$ & 0 (MF) & 0.335$\pm$0.000 & \textbf{0.355$\pm$0.006} & 2.056$\pm$0.000 \\
 & 1 & \textbf{0.323$\pm$0.001} & 0.366$\pm$0.007 & \textbf{2.080$\pm$0.000} \\
 & 2 & \textbf{0.323$\pm$0.000} & 0.364$\pm$0.004 & \textbf{2.080$\pm$0.000} \\ \midrule
$L=3$ & 0 (MF) & 0.163$\pm$0.002 & 0.240$\pm$0.010 & 2.107$\pm$0.000 \\
 & 1 & 0.151$\pm$0.005 & \textbf{0.221$\pm$0.007} & \textbf{2.129$\pm$0.000} \\
 & 2 & \textbf{0.149$\pm$0.003} & 0.226$\pm$0.013 & \textbf{2.129$\pm$0.000} \\ \midrule
$L=5$ & 0 (MF) & 0.184$\pm$0.005 & 0.229$\pm$0.002 & 2.115$\pm$0.000 \\
 & 1 & \textbf{0.151$\pm$0.007} & \textbf{0.206$\pm$0.006} & 2.132$\pm$0.000 \\
 & 2 & 0.153$\pm$0.005 & 0.212$\pm$0.005 & \textbf{2.133$\pm$0.000} \\ \bottomrule
\end{tabular}

\end{sc}
\end{small}
\vskip -0.1in
\end{table*}

\begin{table*}[tbp]
\caption{
Results comparing models trained using different variational distributions.
\gls{DGMRF} models with $L \in \{1,3,5\}$ layers have been trained on the Chameleon Wikipedia data and the California housing data (without features).
All values are reported as mean$\pm$standard deviation across 5 random seeds.
Note that higher \gls{ELBO} values are better.
}
\label{tab:vi_res2}
\vskip 0.15in
\centering
\begin{small}
\begin{sc}
\begin{tabular}{@{}llllcccc@{}}
\toprule
& & \multicolumn{3}{c}{\textbf{Chameleon}} & \multicolumn{3}{c}{\textbf{California Housing, No Features}} \\ 
\cmidrule(lr){3-5} \cmidrule(lr){6-8}
\textbf{DGMRF} & \textbf{VI Layers} & \multicolumn{1}{c}{RMSE} & \multicolumn{1}{c}{CRPS} & ELBO & RMSE ($\times 100$) & CRPS ($\times 100$) & ELBO \\ \midrule
$L=1$ & 0 (MF) & 1.617$\pm$0.001 & 0.899$\pm$0.001 & -1.026$\pm$0.001 & 7.074$\pm$0.001 & 3.743$\pm$0.001 & 0.438$\pm$0.000 \\
 & 1 & \textbf{1.589$\pm$0.000} & \textbf{0.883$\pm$0.000} & -1.000$\pm$0.000 & 6.909$\pm$0.000 & 3.665$\pm$0.000 & 0.488$\pm$0.000 \\
 & 2 & \textbf{1.589$\pm$0.001} & \textbf{0.883$\pm$0.000} & \textbf{-0.999$\pm$0.000} & \textbf{6.897$\pm$0.002} & \textbf{3.661$\pm$0.001} & \textbf{0.490$\pm$0.000} \\ \midrule
$L=3$ & 0 (MF) & 1.527$\pm$0.002 & 0.845$\pm$0.001 & -1.002$\pm$0.001 & 6.914$\pm$0.001 & 3.702$\pm$0.002 & 0.491$\pm$0.000 \\
 & 1 & \textbf{1.511$\pm$0.006} & \textbf{0.835$\pm$0.003} & \textbf{-0.981$\pm$0.008} & \textbf{6.853$\pm$0.000} & 3.656$\pm$0.001 & \textbf{0.500$\pm$0.000} \\
 & 2 & 1.515$\pm$0.014 & 0.837$\pm$0.007 & -0.986$\pm$0.010 & \textbf{6.853$\pm$0.001} & \textbf{3.654$\pm$0.001} & \textbf{0.500$\pm$0.000} \\ \midrule
$L=5$ & 0 (MF) & 1.489$\pm$0.063 & 0.813$\pm$0.039 & -0.987$\pm$0.044 & \multicolumn{3}{c}{-} \\
 & 1 & 1.465$\pm$0.033 & 0.804$\pm$0.023 & -0.963$\pm$0.011 & \multicolumn{3}{c}{-} \\
 & 2 & \textbf{1.453$\pm$0.020} & \textbf{0.795$\pm$0.015} & \textbf{-0.961$\pm$0.009} & \multicolumn{3}{c}{-} \\ \bottomrule
\end{tabular}

\end{sc}
\end{small}
\vskip -0.1in
\end{table*}

%% file: tables/wiki_std.tex
\begin{table*}[tb]
\caption{Standard deviations of \gls{DGMRF} results for the Wikipedia experiment.
Each standard deviation is computed across the results of 5 models trained using different random seeds.}
\label{tab:wiki_std}
\vskip 0.15in
\centering
\begin{small}
\begin{sc}
\begin{tabular}{@{}lccccccc@{}}
\toprule
& \multicolumn{2}{c}{\textbf{Chameleon}} & \multicolumn{2}{c}{\textbf{Squirrel}} & \multicolumn{2}{c}{\textbf{Crocodile}} \\
\cmidrule(lr){2-3} \cmidrule(lr){4-5} \cmidrule(lr){6-7}
\textbf{\gls{DGMRF}} & RMSE & CRPS & RMSE & CRPS & RMSE & CRPS  \\ \midrule
$L=1$ & 0.000 & 0.000 & 0.000 & 0.000 & 0.000 & 0.000 \\
$L=3$ & 0.006 & 0.003 & 0.034 & 0.021 & 0.031 & 0.018 \\
$L=5$ & 0.033 & 0.023 & 0.022 & 0.013 & 0.030 & 0.019 \\ \bottomrule
\end{tabular}
\end{sc}
\end{small}
\vskip -0.1in
\end{table*}

%% file: tables/spatial_std.tex
\begin{table}[tb]
\caption{Standard deviations of \gls{DGMRF} results for the the California housing and wind speed data.
Each standard deviation is computed across the results of 5 models trained using different random seeds.
Values for the California housing data are multiplied by a factor 100 to match the scale in \reftab{cal_res}.}
\label{tab:spatial_std}
\vskip 0.15in
\centering
\begin{small}
\begin{sc}
\begin{tabular}{@{}lcccccccc@{}}
\toprule
& \multicolumn{2}{c}{\textbf{Cal., No Features}} & \multicolumn{2}{c}{\textbf{Cal., Features}} & \multicolumn{2}{c}{\textbf{Wind, Spatial Mask}} & \multicolumn{2}{c}{\textbf{Wind, Random Mask}} \\
 \cmidrule(lr){2-3} \cmidrule(lr){4-5} \cmidrule(lr){6-7} \cmidrule(lr){8-9}
\textbf{\gls{DGMRF}} & RMSE & CRPS & RMSE & CRPS & RMSE & CRPS & RMSE & CRPS\\ \midrule
$L=1$ & 0.000 & 0.000 & 0.001 & 0.001 & 0.013 & 0.008 & 0.000 & 0.000\\ 
$L=2$ & 0.000 & 0.001 & 0.001 & 0.001 & 0.011 & 0.007 & 0.001 & 0.000\\ 
$L=3$ & 0.000 & 0.001 & 0.001 & 0.001 & 0.017 & 0.011 & 0.001 & 0.000\\ \bottomrule
\end{tabular}
\end{sc}
\end{small}
\vskip -0.1in
\end{table}

%% file: appendices/reparametrization.tex
When parametrizing the model we want to avoid the situation where $|\alpha_l| = |\beta_l|$, as this would lead to a $\Glayer{l}$ with determinant 0 and a non-definite precision matrix $Q$.
Even if this exact situation would likely be avoided during optimization, this represents an unstable area of the parameter space as the log-determinant goes towards $-\infty$.
We solve this problem by enforcing the inequality $|\alpha_l| > |\beta_l| \Leftrightarrow \left|\frac{\beta_l}{\alpha_l}\right| < 1$.
We choose this direction of the inequality as it will guarantee the convergence of the power series that we use in one of our methods for the log-determinant computation (see \refapp{power_series_details}).
It also makes intuitive sense that the central node should have a larger influence in each layer than it's neighbors.
We also parametrize the model so that $\alpha_l > 0$, which is no restriction on the model class.
See this for example in how \refeq{dgmrf_def} is invariant to the sign of the right hand side.

To achieve these inequalities, as well as $\gamma_l \in \mathopen]0,1\mathclose[$, we reparametrize the model as 
\al[reparametrization]{
    \alpha_l &= \exp\left(\thetal{l}_1\right)\\
    \beta_l &= \alpha_l \tanh\left(\thetal{l}_2\right)\\
    \gamma_l &= \sigmoid\left(\thetal{l}_3\right)
}
where $\thetal{l}_1, \thetal{l}_2, \thetal{l}_3 \in \R$ are free parameters.
This allows us to use unconstrained optimization while preserving the desired restrictions on $\Glayer{l}$.
In a similar way we reparametrize $\sigma = \exp\left(\theta_\sigma \right) > 0$ with a free parameter $\theta_\sigma \in \R$.

%% file: appendices/power_series.tex
To make our graph DGMRF scale to large graphs the computation of $\logdet{\Glayer{l}}$ has to be both efficient and differentiable w.r.t.\ all layer parameters.
We follow the approach of \citeauth{iresnet}, where a stochastic approximation of the log-determinant is constructed based on a power series.

We start by rewriting $\Glayer{l}$ as
\als[G_ps_rewrite]{
    \Glayer{l} &= D^{\gamma_l}(\alpha_l I + \beta_l D^{-1}A)\\
    &= D^{\gamma_l - \frac{1}{2}} (\alpha_l D^{\frac{1}{2}}D^{-\frac{1}{2}} + \beta_l D^{-\frac{1}{2}} A D^{-\frac{1}{2}})D^{\frac{1}{2}}\\
    &= \alpha_l D^{\gamma_l - \frac{1}{2}} \left( I + \frac{\beta_l}{\alpha_l} \tilde{A}\right)D^{\frac{1}{2}}
}
where we again use the definition $\tilde{A} = D^{-\frac{1}{2}} A D^{-\frac{1}{2}}$. 
Then
\als[G_ps_logdet]{
    \logdet{\Glayer{l}} &= \log\left(\left|\det\left(\alpha_l D^{\gamma_l - \frac{1}{2}} \left( I + \frac{\beta_l}{\alpha_l}\tilde{A}\right)D^{\frac{1}{2}}\right)\right|\right)\\
    &= \log\left(\left|\alpha_l^N \det\left(D\right)^{\gamma_l - \frac{1}{2}} \det\left( I + \frac{\beta_l}{\alpha_l}\tilde{A}\right)\det\left(D\right)^{\frac{1}{2}}\right|\right)\\
    &= N\log\left(\alpha_l\right) + \gamma_l\log\left(\det\left(D\right)\right) + \logdet{I + \frac{\beta_l}{\alpha_l}\tilde{A}}
}
where we have used that $\alpha_l > 0$ and $\det(D) > 0$ as well as a number of properties of determinants (see for example \citetapp{matrixcookbook}).
All involved quantities are simple and efficient to compute, except for $\logdet{I + \frac{\beta_l}{\alpha_l}\tilde{A}}$ which involves the adjacency matrix of the graph.
To tackle this challenge, we will formulate this quantity as a power series and apply a stochastic approximation.

We first note a few things about the matrix $\tilde{A}$. 
Firstly, the eigenvalues of $\tilde{A}$ lie in $[-1,1]$. 
This follows for example from the fact that the normalized graph Laplacian $I - \tilde{A}$ has eigenvalues in $[0,2]$ (\citealp{graphsandspectra}; \citealpapp{spectra_of_weighted_graphs}). %
Secondly, since $\tilde{A}$ is symmetric we can also conclude that its singular values are given by the absolute values of its eigenvalues\footnote{Let $\b{v}_i$ and $\lambda_i$ be the $i$:th eigenvector and eigenvalue of a symmetric matrix $B$. Then $B^\transpose B \b{v}_i = B^2 \b{v}_i = \lambda_i B \b{v}_i = \lambda_i^2 \b{v}_i$ and $\sqrt{\lambda_i^2} = |\lambda_i|$ is a singular value of $B$.}.
In particular, the maximum singular value $\sigma_{\max}(\tilde{A}) \leq 1$.

To construct our approximation, we first show that $\det\left(I + \frac{\beta_l}{\alpha_l}\tilde{A}\right) > 0$.
Let $\tilde{\lambda}_i \in [-1, 1]$ be an eigenvalue of $\tilde{A}$ with eigenvector $\b{w}_i$. 
Then
\eq[Atildeeig]{
    \left(I + \frac{\beta_l}{\alpha_l}\tilde{A}\right)\b{w}_i = \b{w}_i + \frac{\beta_l}{\alpha_l}\tilde{A} \b{w}_i = \left(1 + \frac{\beta_l}{\alpha_l} \tilde{\lambda}_i \right) \b{w}_i,
}
so $\psi_i = 1 + \frac{\beta_l}{\alpha_l} \tilde{\lambda}_i$ is an eigenvalue to $I + \frac{\beta_l}{\alpha_l}\tilde{A}$. 
By our parametrization $\frac{\beta_l}{\alpha_l} \in \mathopen]-1,1\mathclose[$\,, which implies that $\psi_i > 0 \, \forall i$. 
Hence 
\eq[posdet]{
\det\left(I + \frac{\beta_l}{\alpha_l}\tilde{A}\right) = \prod_{i=1}^N \psi_i > 0 \Rightarrow
\logdet{I + \frac{\beta_l}{\alpha_l}\tilde{A}} = 
\log\left(\det\left( I + \frac{\beta_l}{\alpha_l}\tilde{A}\right)\right).
}
Similarly to \citeauth{iresnet} we next formulate the log-determinant in terms of the power series
\eq[powerseries]{
    \log\left(\det\left( I + \frac{\beta_l}{\alpha_l}\tilde{A}\right)\right) = 
    \trace\left(\log\left( I + \frac{\beta_l}{\alpha_l}\tilde{A}\right)\right) = 
    \sum_{k=1}^\infty (-1)^{k+1} \frac{\trace\left(\left(\frac{\beta_l}{\alpha_l}\tilde{A}\right)^k\right)}{k} 
    =
    \sum_{k=1}^\infty - \frac{1}{k} \left(-\frac{\beta_l}{\alpha_l}\right)^k \trace\left(\tilde{A}^k \right).
}
Using the properties of $\tilde{A}$ discussed earlier we can show that 
\eq[series_converge]{
\norm{\frac{\beta_l}{\alpha_l}\tilde{A}} = \paramq \norm{\tilde{A}} = \paramq \sigma_{\max}\left(\tilde{A} \right) < 1
}
which implies that the power series in \refeq{powerseries} converges.
We can then truncate this series at some large $k=K$ to get an approximation, as discussed in \refsec{power_series}.

\begin{figure}[tb]
\centering
\includegraphics[width=0.8\columnwidth]{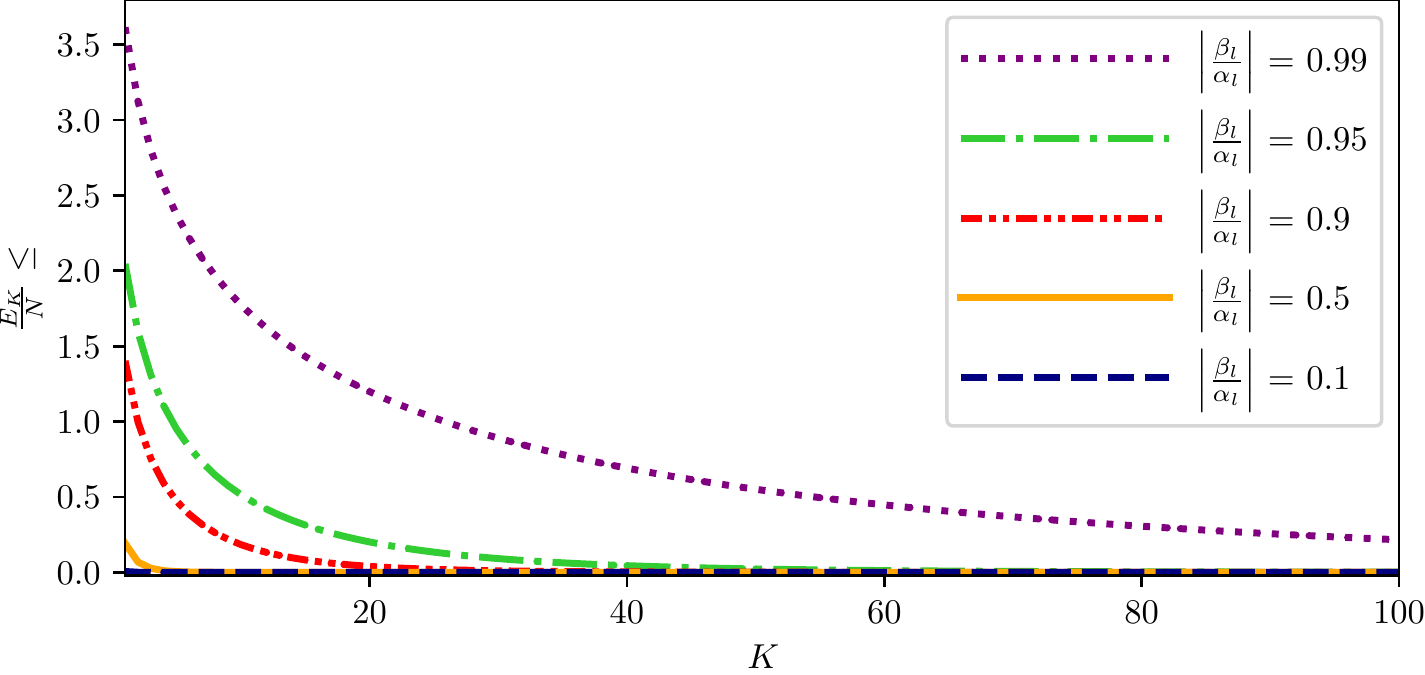}
\caption{The quantity $-\log\left(1 - \paramq \right) - \sum_{k=1}^{K} \frac{1}{k} \paramq^k$ from the bound in \refeq{power_series_bound}, plotted as a function of $K$ for different parameter values.}
\label{fig:power_series_bound}
\end{figure}

Using the same reasoning as \citeauth{iresnet}, the truncation error of the series can be bounded.
We note that
\eq[single_trace_bound]{
    \left|\trace\left(\tilde{A}^k \right)\right| = \left|\sum_{i=1}^N \tilde{\lambda}_i^k\right| \leq \sum_{i=1}^N \left|\tilde{\lambda}_i^k\right| \leq N
}
where we have used that each $\tilde{\lambda}_i^k$ is an eigenvalue to $\tilde{A}^k$ and that $\left|\tilde{\lambda}_i^k\right| \leq 1$.
The absoute error $E_K$ of the series in \refeq{powerseries} truncated at $K$ can then be bounded as
\eq[power_series_bound]{
    E_K \leq N \left(-\log\left(1 - \paramq \right) - \sum_{k=1}^{K} \frac{1}{k} \paramq^k \right).
}
We refer to \citeauth{iresnet} for a proof of this bound. 
Our case is fully analogues and can be derived by replacing their Lipschitz constant by $\paramq$.
For the majority of parameter values this bound decreases very quickly with $K$, as shown in \reffig{power_series_bound}.
It is only when $\paramq$ is close to 1 that the rate of decrease slows down.

As the traces in \refeq{powerseries} only depend on the graph structure, their values can be pre-computed and stored for use during the model training.
Computing $\trace\left(\tilde{A}^k \right)$ can for example be done using sparse linear algebra, multiplying together $k$ sparse matrices.
For large values of $k$ these matrix products do however become increasingly dense and the computation inefficient.
An alternative approach is to use the fact that
\eq[stochastic_trace]{
    \trace\left( \tilde{A}^k \right) = \E{p(\b{u})}{\b{u}^\transpose \tilde{A}^k\b{u}}
}
for any distribution $p(\b{u})$ with zero mean and covariance matrix $I$ \citeapp{hutchinson_estimator}.
Taking a \gls{MC} estimate of \refeq{stochastic_trace} yields the Hutchinson's trace estimator.
We note that while the estimator is unbiased, we here only compute its value during pre-processing and then reuse it during training.
This gives an approximation of the trace, but since we do not have to re-compute it during training we can use a high value for $K$ and many samples for the \gls{MC} estimate.
We choose $p(\b{u})$ as the uniform distribution over $\{-1, 1\}^\nnodes$, resulting in a minimum variance estimator.
Computing this estimate only requires applying $\tilde{A}$ to vectors sampled from $p(\b{u})$.
As the multiplication with $\tilde{A}$ corresponds to a single \gls{GNN} layer we implement this using the same \gls{GNN} framework as the rest of the model.

%% file: appendices/baselines.tex
In this appendix we describe the baseline models used in our experiments.

\subsection{\acrfull{LP}}
The label propagation baseline follows the basic framework of \cite{lp}, but using real-valued labels \cite{residual_corr_gnns}.
This approach boils down to finding a joint solution to the equation
\eq[lp_harmonic]{
    y_i = \frac{1}{d_i} \sum_{j \in \neigh{i}} y_j
}
for all $i$ corresponding to unobserved nodes.
Such a solution exists, but requires the inversion of a potentially very large matrix. 
We utilize the \gls{CG} method for this.

The \gls{LP} baseline does not give a predictive distribution that can be used for computing \gls{CRPS}.
As there is no randomness in the method we can not apply the ensemble method for uncertainty estimation here.

\subsection{\acrfull{IGMRF}}
The \gls{IGMRF} baseline is a \gls{GMRF} with mean $\b{\mu} = \b{0}$ and precision matrix $Q = \kappa (D - A) + \epsilon I$.
Here $\kappa$ is a parameter of the model and $\epsilon$ a small value added to the diagonal in order to ensure positive definiteness and numerical stability.
We use $\epsilon = 10^{-4}$ for the Wikipedia graphs and $\epsilon = 10^{-6}$ for all other experiments.
Similarly to our \gls{DGMRF}, this \gls{GMRF} defines a prior over $\x$ and we then use a Gaussian likelihood with noise variance $\sigma^2$.
The parameters $\kappa$ and $\sigma$ are optimized by maximizing the log marginal likelihood $\log p(\y_m)$.
For this we use a simple grid-based optimization where $\log p(\y_m)$ is evaluated for each combination of $\sigma \in \{0.001, 0.01, 0.1, 1\}$ and 20 log-spaced values of $\kappa$ in the range $[0.01, 1000]$. 
To compute the log marginal likelihood we use the fact that
\eq[lml_trick]{
    p(\y_m) = \left.\frac{p(\y_m|\x)p(\x)}{p(\x|\y_m)}\right\vert_{\x = \x'}
}
holds for any $\x'$.
We then utilize sparse linear algebra computations, such as the sparse Cholesky decomposition, to compute all quantities involved \cite{gmrf_book}.
In contrast to our \gls{DGMRF} this is feasible here since $Q$ is very sparse.

\subsection{Graph GP}
For the Graph \gls{GP} baseline we use the reference implementation made available\footnote{\url{https://github.com/spbu-math-cs/Graph-Gaussian-Processes}} by the authors \cite{matern_graph_gp}.
In order to make the model scale to large graphs we use a number of approximation techniques, all described in the original paper.
For training the model we use doubly stochastic variational inference with a batch size of 128 and the proposed interdomain inducing variables.
The kernel matrix is approximated using the 500 smallest eigenvalues of the graph Laplacian and corresponding eigenvectors.
When applicable we use the weighted adjacency matrix also for the Graph \gls{GP} baseline.

\subsection{\acrfull{SVGP}}
The \gls{SVGP} \cite{svgp} implementation is from the GPyTorch library \citeapp{gpytorch}.
We use a Matérn kernel with $\nu = 3/2$, automatic relevance determination and a batch size of 256.
The positions of 500 inducing points are learned jointly with the model parameters.
A Gaussian variational distribution is used, where we parametrize the Cholesky factor of the covariance matrix directly.
When more features than the spatial coordinates are included, the features and coordinates are concatenated and together used as kernel input.

\subsection{\acrfull{Bayes LR}}
The Bayesian linear regression model is a default Bayesian Ridge model from the scikit-learn library \citeapp{scikit-learn}.
It uses a Gaussian prior on the regression coefficients and uninformative gamma priors on regularization parameters.

\subsection{\acrfull{MLP}}
In comparison to other baseline models, the deep learning baselines require some tuning of the network architectures.
To achieve this we reserve 20\% of the training data (observed nodes) as a validation set. 
No such validation set is used for \glspl{DGMRF} or other baseline models.
The validation set is used to decide the network architecture and for regularization by early stopping.

For the \gls{MLP} baseline we consider the layer configurations $(\text{number of hidden layers} \times \text{dimensionality}) \in \{
1 \times 128,\linebreak[0] 
1 \times 512,\linebreak[0] 
2 \times 128,\linebreak[0] 
2 \times 512
\}$.
We use an MSE loss and the Adam optimizer \citeapp{adam} to train one model of each configuration.
The layer configuration resulting in the lowest validation loss is then used for the ensemble.

\subsection{\glspl{GNN} (\acrfull{GCN} and \acrfull{GAT})}
The \gls{GNN} models are trained in the same way as the \gls{MLP}, also using 20\% of the observed nodes for validation.
The full graph is fed to the \gls{GNN} at each iteration and the MSE loss computed for all nodes designated for training.
We here consider the layer configurations $(\text{number of hidden layers} \times \text{dimensionality}) \in \{
1 \times 32,\linebreak[0]
1 \times 64,\linebreak[0]
3 \times 32,\linebreak[0] 
3 \times 64,\linebreak[0] 
3 \times 128,\linebreak[0] 
5 \times 32,\linebreak[0] 
5 \times 64,\linebreak[0] 
5 \times 128,\linebreak[0] 
7 \times 128 
\}$.
We experiment with two \gls{GNN} baselines, the \acrfull{GCN} of \citeauth{gcn} and the \acrfull{GAT} of \citeauth{gat}. 
For the \gls{GAT} model we use single-head attention.

\subsection{\acrfull{DGP}}
As earlier noted the \gls{DGMRF} model can be related to the message passing formulation of \glspl{GNN}.
An interesting question arises about how much of the performance of these types of models can be attributed to any inductive biases of the \gls{GNN} layer formulation.
We ask, does the \gls{GNN} architecture in itself restrict the model class to a set of models that are highly useful for common types of signals on graphs?
\citeauth{dip} investigate a similar question for \glspl{CNN} and natural images in their Deep Image Prior model.
For the in-painting setting, random noise is fed to the network and the model then trained on the observed pixels of a single picture.
The idea is that any inductive biases in the model architecture can still steer the training to good solutions.
As the model is massively overparametrized the training should not be allowed to converge, but stopped at some earlier point that still results in a good solution.
We hypothesize that \glspl{GNN} can exhibit similar properties as the \glspl{CNN} investigated by \citeauth{dip} and implement a graph version that we denote \gls{DGP}.

For the \gls{DGP} model we use the same setup for training \glspl{GNN} as described above.
We use \gls{GCN} layers, tune the layer configuration, and use early stopping in the same way as before.
A single Gaussian sample is drawn and used as the model input throughout the whole training.

The \gls{DGP} model allows for applying \glspl{GNN} to graphs without input features.
As \gls{DGP} has proven useful in some of our experiments this indicates that our earlier stated hypothesis has some ground to it.
Exploring this type of model and the inductive biases of \glspl{GNN} in general is an interesting direction for further research.

%% file: appendices/experiment_details.tex
In the interest of reproducability we present additional details about the experiment setups and datasets in this appendix.
For training our \gls{DGMRF} we use a learning rate of 0.01 and the Adam optimizer \citeapp{adam} in all experiments.
The model has not been observed to be sensitive to these choices so no extensive tuning has been done.
Note that overfitting is not a considerable problem here.
If necessary the learning rate can be tuned to make the \gls{ELBO} converge, just using the training data (observed nodes).
On synthetic data we train the model for 50\,000 iterations, on the Wikipedia and California housing datasets 80\,000 iterations (150\,000 for 5-layer \glspl{DGMRF}) and for the wind speed data 150\,000 iterations.
These numbers are large enough for the \gls{ELBO} to converge and often unnecessarily high, meaning that runtimes could be slightly reduced with a more carfeful choice.
In all experiments we use one \gls{DGMRF} layer for $\tilde{G}$ in the variational distribution $q$ (see \refeq{vi_layers}).
At each iteration of training we draw 10 samples from $q$ to estimate the expectation in the \gls{ELBO}.

An overview of the different graphs used in experiments is presented in \reftab{graph_details}.
\input{tables/dataset_details}

\subsection{\synthexp}
\label{sec:synthexp_details}
\subsubsection{Synthetic \gls{DGMRF} data}
The synthetic data used for the experiment in \reffig{layer_results_mean} was sampled from 1--4 layer \glspl{DGMRF} based on a synthetic generated graph.
A graph with 3000 nodes was generated by first sampling node positions uniformly over $[0,1]^2$ and then forming the graph using a Delaunay triangulation \cite{triangulations_book}.
The positions of the nodes were only used to construct the graph and for visualization purposes.
\glspl{DGMRF} with parameters 
$\alpha_l = 1.2$,
$\beta_l = -1.0$,
$\gamma_l = 1$,
and $b_l = 0$
for all layers were then defined on this graph.
The final data $\y$ was constructed by drawing a single sample $\x$ from a \gls{DGMRF} and adding Gaussian noise with $\sigma = 0.01$.
Observation masks were additionally created by randomly setting 25\% of nodes as unobserved.

\subsubsection{Dense}
For the synthetic Dense data (discussed in \refapp{synth_exp_extra}), a random graph was generated in a similar way as above.
Based on this random graph the corresponding 3-hop graph was then created.
A sample $\x$ was drawn from a 1-layer \gls{DGMRF} with
$\alpha_l = 1.2$,
$\beta_l = -1.0$,
$\gamma_l = 1$,
and $b_l = 0$
defined on this 3-hop graph.
Gaussian noise with $\sigma = 0.01$ was again added.

\subsubsection{Mix}
For the Mix data a graph with 5000 nodes was created in the same way as above.
A \gls{GMRF} was defined on this graph with mean $\b{\mu} = \b{0}$ and precision matrix $Q = \sum_{i=1}^4 G_i^\transpose G_i$, where each $G_i$ was defined as a \gls{DGMRF} layer according to \refeq{layerG}.
We defined each $G_i$ not directly on the generated graph, but on the $i$-hop graph $\G^i$ (in the same way as the Dense dataset).
The parameters of each $G_i$ were sampled randomly as
$\alpha_l \sim \uniform{[0.5, 1.5]}$,
$\beta_l \sim \uniform{[-1.1, -0.1]}$
and 
$\gamma_l$ fixed to $1$.
The data was then sampled in the same way as above.

\subsection{\wikiexp}
The Wikipedia graphs were created and made available\footnote{\url{https://github.com/benedekrozemberczki/MUSAE}} by \citeauth{wikipedia_datasets}.
In our experiments we do not use the accompanying node features and we use the logarithm of target values.
We pre-processed the data by removing duplicate edges and self-loops (edges with the same node as both endpoints).

After tuning the \gls{DGP} model the architecture used for the ensemble was $5 \times 128$ on the Crocodile data and $7 \times 128$ on the Chameleon and Squirrel data.

\subsection{\calexp}
The California housing dataset was loaded directly through the scikit-learn library\footnote{\url{https://scikit-learn.org/stable/datasets/real_world.html\#california-housing-dataset}} \citeapp{scikit-learn}.
We used the logarithm of target values and performed the same feature transformation as described in the original paper \cite{california}.
Outliers in the data were removed by the elliptic envelope method, determining the contamination degree by visual inspection.
Features and targets were finally normalized to $[0,1]$.

An equirectangular projection was used on the latitude and longitude of each housing block to create the node positions.
Based on these positions the graph was created as a Delaunay triangulation.
The graph creation was done using the built in Delaunay computation in PyTorch Geometric \cite{pytorch_geometric}, which in turn relies on the Qhull library\footnote{\url{http://www.qhull.org/}}.
We note that many choices exist for how to create such a spatial graph, but the Delaunay method has in our experiments shown to work well while maintaining a highly sparse graph.
Considering different algorithms for graph creation is outside the scope of this work.
The graph was weighted with $w_{i,j} = (\rho_{i,j} + \epsilon)^{-1}$ based on the euclidean distance $\rho_{i,j}$ between nodes $i$ and $j$.
A small $\epsilon$ is included to prevent division by zero for nodes assigned to the same position.

For handling features in the \gls{DGMRF} model we follow the approach of \citeauth{dgmrf} and use an auxiliary Bayesian linear regression model.
An uninformative $\normal{\b{0}}{10^{8} I}$ prior and a mean-field variational distribution are used for the coefficients of the linear model.
For posterior inference the coefficients can be integrated out.

When no features are used the baseline models in \reftab{cal_res} only utilize the node positions as inputs.
When features are used both the node positions and the pre-processed socio-economic features are used.
The Graph \gls{GP}, \gls{LP} and \gls{IGMRF} baselines only utilize the graph and no node positions or other features.
There is no obvious way to adapt these methods to also use node features, so this was deemed outside the scope of this paper.

The resulting layer configurations after tuning for the deep learning baselines are listed in \reftab{dl_params} together with the number of trainable parameters. 
These numbers can be compared to the significantly smaller $4L+1$ trainable parameters in an $L$-layer \gls{DGMRF}.

\input{tables/spatial_params}

\subsection{\windexp}
The wind speed data originates from the Wind Integration National Dataset Toolkit\footnote{\url{https://www.nrel.gov/grid/wind-toolkit.html}}.
The mean wind speed in the summary statistics for the different sites was used as our target.
To somewhat limit ourselves to continental US we only included sites at a latitude $\geq 24.4^{\circ}$.
Also here an equirectangular projection was used and the graph then created in the same way as for the California housing data.

The \gls{LP} and \gls{IGMRF} baseline models only utilize the graph in this experiment.
All other baselines (not the \glspl{DGMRF}) use the node positions as features, but here expanded to second degree polynomial features in order to capture some non-linear trends.
Deep learning architectures are listed in \reftab{dl_params}.
We were not able to apply the Graph \gls{GP} to this dataset as the required pre-computation of eigenvalues and eigenvectors does not scale to a graph of this size.

For the \gls{DGMRF} we here use the power series method of \refsec{power_series} for the log-determinant computations.
We truncate the power series in \refeq{power_series} at $K = 50$ terms and use the Hutchinson's trace estimator with 1000 \gls{MC} samples (see \refapp{power_series_details}).

\subsection{\obsexp}
In this experiment we only changed the mask specifying which nodes were observed in the Mix and Wikipedia Crocodile graphs.
Such masks were created for 5\%, 20\%, 40\%, 60\%, 80\% and 95\% of nodes being observed.
Each set of observed nodes was chosen to be a superset of the previous one, with the additional nodes sampled uniformly at random.

In this experiment we use 5 random seeds also for the \gls{DGP} baseline, in order to compute confidence intervals.
This means that the whole ensemble of 10 models is re-trained for each random seed, in total creating 50 models for each observation mask.
To make this feasible we fix the \gls{GNN} architecture used.
Based on a preliminary tuning experiment we found good choices to be the $7 \times 128$ architecture for the Mix data and the $5 \times 32$ architecture for the Wikipedia Crocodile data.

%% file: tables/dataset_details.tex
\begin{table}[tb]
\caption{Properties of the different graph datasets. The density of an undirected graph with $N$ nodes and $N_e$ edges is computed as $2 N_e / {N(N-1)}$. 
This is equal to the fraction of possible edges present, compared to the complete graph.
The (unweighted) diameter of a graph is the longest shortest path between any two nodes in the graph.
}
\label{tab:graph_details}
\vskip 0.15in
\centering
\begin{small}
\begin{sc}
\begin{tabular}{@{}lrrccc@{}}
\toprule
\textbf{Graph} & \textbf{Nodes} & \textbf{Edges} & \textbf{Density} & \textbf{Diameter} & \textbf{Unobserved nodes} \\ \midrule
Synthetic DGMRF & 3\,000 & 8\,977 & 0.0020 & 32 & 25\% \\
Dense & 3\,000 & 8\,974 & 0.0020 & 32 & 25\% \\
Mix & 5\,000 & 14\,970 & 0.0012& 41 & 50\% \\
Wikipedia chameleon & 2\,277 & 31\,371 & 0.0121 & 11 & 50\% \\
Wikipedia squirrel & 5\,201 & 198\,353 & 0.0147 & 10 & 50\% \\
Wikipedia crocodile & 11\,631 & 170\,773 & 0.0025 & 11 & 50\% \\
California housing & 20\,536 & 61\,585 & $2.92\times 10^{-4}$ & 57 & 50\% \\
Wind Speed, spatial mask & 126\,652 & 379\,933 & $4.74\times 10^{-5}$ & 101 & 9.12\% \\
Wind speed, random mask & 126\,652 & 379\,933 & $4.74\times 10^{-5}$ & 101 & 50\% \\ \bottomrule
\end{tabular}
\end{sc}
\end{small}
\vskip -0.1in
\end{table}

%% file: tables/spatial_params.tex
\begin{table}[tb]
\caption{Layer configurations (number of hidden layers $\times$ dimensionality) and corresponding number of trainable parameters of the deep learning models. Each layer configuration is the final one decided on from hyperparameter tuning on the validation data.}
\label{tab:dl_params}
\vskip 0.15in
\centering
\begin{small}
\begin{sc}
\begin{tabular}{@{}llclclclc@{}}
\toprule
 & \multicolumn{2}{c}{\textbf{Cal., No Features}} & \multicolumn{2}{c}{\textbf{Cal., Features}} & \multicolumn{2}{c}{\textbf{Wind, Spatial Mask}} & \multicolumn{2}{c}{\textbf{Wind, Random Mask}} \\ 
 \cmidrule(lr){2-3} \cmidrule(lr){4-5} \cmidrule(lr){6-7} \cmidrule(lr){8-9}
\textbf{Model} & \multicolumn{1}{c}{Layers} & \# Param. & \multicolumn{1}{c}{Layers} & \# Param. & \multicolumn{1}{c}{Layers} & \# Param. & \multicolumn{1}{c}{Layers} & \# Param. \\ \midrule
MLP & $2 \times 512$ & 264\,705 & $2 \times 512$ & 268\,801 & $2 \times 512$ & 266\,241 & $2 \times 512$ & 266\,241 \\
GCN & $7 \times 128$ & 99\,585 & $3 \times 64$ & 9\,089 & $7 \times 128$ & 99\,969 & $7 \times 128$ & 99\,969 \\
GAT & $5 \times 128$ & 67\,841 & $3 \times 128$ & 35\,329 & $5 \times 32$ & 4\,769 & $5 \times 64$ & 17\,729 \\ \bottomrule
\end{tabular}
\end{sc}
\end{small}
\vskip -0.1in
\end{table}